\documentclass[10pt,twocolumn,letterpaper]{article}

\usepackage[pagenumbers]{wacv} 

\usepackage{graphicx}
\usepackage{xcolor}
\usepackage{amsmath, amsthm, amssymb, cases, multirow}
\usepackage{booktabs}
\usepackage{algorithmic}
\usepackage{algorithm}
\usepackage{multirow}
\newcommand{\red}[1]{\textcolor{red}{#1}}
\newcommand{\blue}[1]{\textcolor{blue}{#1}}
\newcommand{\argmax}{\mathop{\rm arg~max}\limits}
\usepackage{enumitem}
\usepackage[accsupp]{axessibility}
\setlist[itemize]{noitemsep, topsep=1pt, parsep=1pt, partopsep=1pt}
\usepackage{here}

\usepackage[pagebackref,breaklinks,colorlinks]{hyperref}

\usepackage[capitalize]{cleveref}
\crefname{section}{Sec.}{Secs.}
\Crefname{section}{Section}{Sections}
\Crefname{table}{Table}{Tables}
\crefname{table}{Tab.}{Tabs.}
\Crefname{equation}{Equation}{Equations}
\crefname{equation}{Eq.}{Eqs.}

\def\wacvPaperID{425} 
\def\confName{WACV}
\def\confYear{2024}

\begin{document}
\title{Active Transfer Learning for Efficient Video-Specific Human Pose Estimation}
\author{
    Hiromu Taketsugu \qquad
    Norimichi Ukita \\
    Toyota Technological Institute \\
    Nagoya, Japan \\
    {\tt \{sd23426,ukita\}@toyota-ti.ac.jp} \\
    \url{https://github.com/ImIntheMiddle/VATL4Pose-WACV2024}
}
\maketitle

\begin{abstract}
    Human Pose (HP) estimation is actively researched because of its wide range of applications. However, even estimators pre-trained on large datasets may not perform satisfactorily due to a domain gap between the training and test data. To address this issue, we present our approach combining Active Learning (AL) and Transfer Learning (TL) to adapt HP estimators to individual video domains efficiently. For efficient learning, our approach quantifies (i) the estimation uncertainty based on the temporal changes in the estimated heatmaps and (ii) the unnaturalness in the estimated full-body HPs. These quantified criteria are then effectively combined with the state-of-the-art representativeness criterion to select uncertain and diverse samples for efficient HP estimator learning. Furthermore, we reconsider the existing Active Transfer Learning (ATL) method to introduce novel ideas related to the retraining methods and Stopping Criteria (SC). Experimental results demonstrate that our method enhances learning efficiency and outperforms comparative methods.
\end{abstract}

\section{Introduction}
\label{sec:intro}

Analyzing Human Poses (HP) in videos has extensive applications in areas such as security~\cite{HPE_in_Surveillance, pose_anomaly, pose_surveillance}, sports analysis~\cite{HPE_in_succer_players, pose_cvsports, pose_icsports}, healthcare~\cite{HPE_in_bed, HPE_in_healthcare}, computer-aided diagnostics~\cite{ASD1, ASD2}, and performance capture~\cite{DBLP:conf/mva/MatsumotoSMMMMU19, DBLP:conf/eccv/UkitaTK08, DBLP:conf/iccv/UkitaHK09}. In these applications, a massive amount of HPs in videos are necessary. To collect such many HPs, HP estimation~\cite{AlphaPose, HRNet_forHPE, UniPose, DCPose,  OpenPose, Openpifpaf} plays a crucial role, as manually annotating every HP in videos is impractical.

Despite huge training datasets, inaccurate HPs may be observed in the test phase due to a domain gap between the training and test datasets~\cite{HPE_Testtime}.
Table~\ref{tab:domain_gap} shows two examples of such a domain gap.
In both two combinations of HP estimators and datasets, a large performance drop is observed.
This paper addresses this challenge by enhancing the performance of the HP estimator through test-time adaptation.
Specifically, we aim to efficiently adapt a pre-trained HP estimator to each video domain with minimal annotation cost.

\begin{table}[t]
  \caption{Performance degradation in the pose estimation accuracy (\%) due to a domain gap between the training and test datasets.}
  \label{tab:domain_gap}
  \vspace*{-2mm}
    \begin{center}
        \begin{tabular}{l|l| c c}
        \hline
        Method & Dataset & Train & Test\\ \hline \hline
        FastPose~\cite{AlphaPose} & JRDB-Pose~\cite{JRDB-Pose} & 95.31 & 39.50 \\ \hline
        SimpleBaseline~\cite{SimpleBaseline} & PoseTrack21~\cite{PoseTrack21} & 98.95 & 75.50 \\ \hline
        \end{tabular}
    \end{center}
\end{table}

\begin{figure}[t]
  \centering
  \includegraphics[width=\columnwidth]{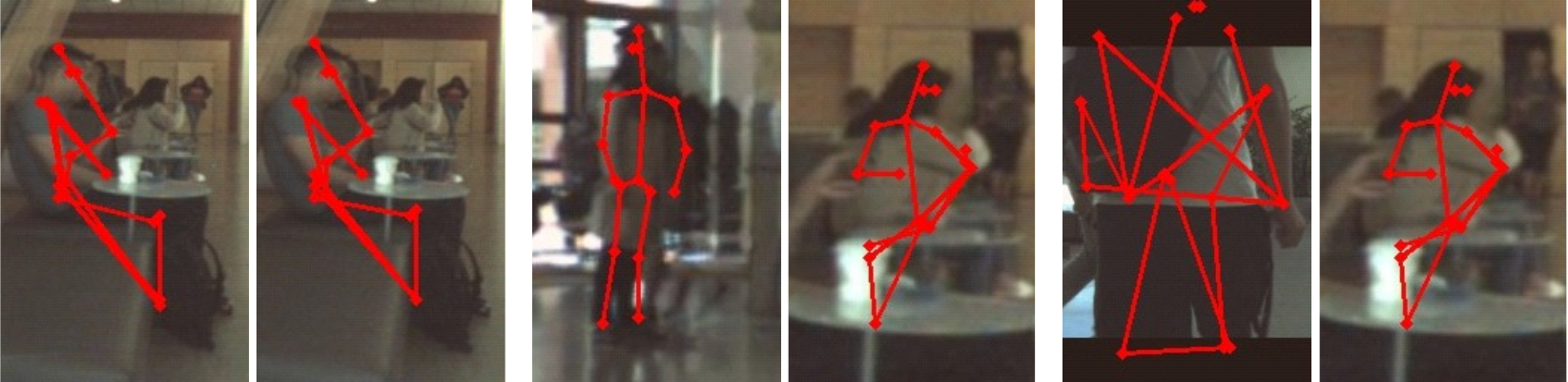}
  \begin{subfigure}{0.32\columnwidth}
    \caption{MPE~\cite{AL_for_HPE}}
    \label{fig:selected_mpe}
  \end{subfigure}
  \begin{subfigure}{0.32\columnwidth}
    \caption{Core-set~\cite{coreset}}
    \label{fig:selected_coreset}
  \end{subfigure}
  \begin{subfigure}{0.32\columnwidth}
    \caption{Ours}
    \label{fig:selected_ours}
  \end{subfigure}
  \vspace*{-2mm}
  \caption{Samples with top scores in each selection criterion. Whereas (a) an uncertainty criterion (MPE~\cite{AL_for_HPE}) selects uncertain but similar samples and (b) a representativeness criterion (Core-Set~\cite{coreset}) selects diverse but uninformative samples, (c) our criteria (THC+WPU+DUW) selects uncertain and diverse samples.}
  \label{fig:selected_examples}
\end{figure}

\begin{figure*}[t]
  \begin{center}
    \includegraphics[width=0.8\linewidth]{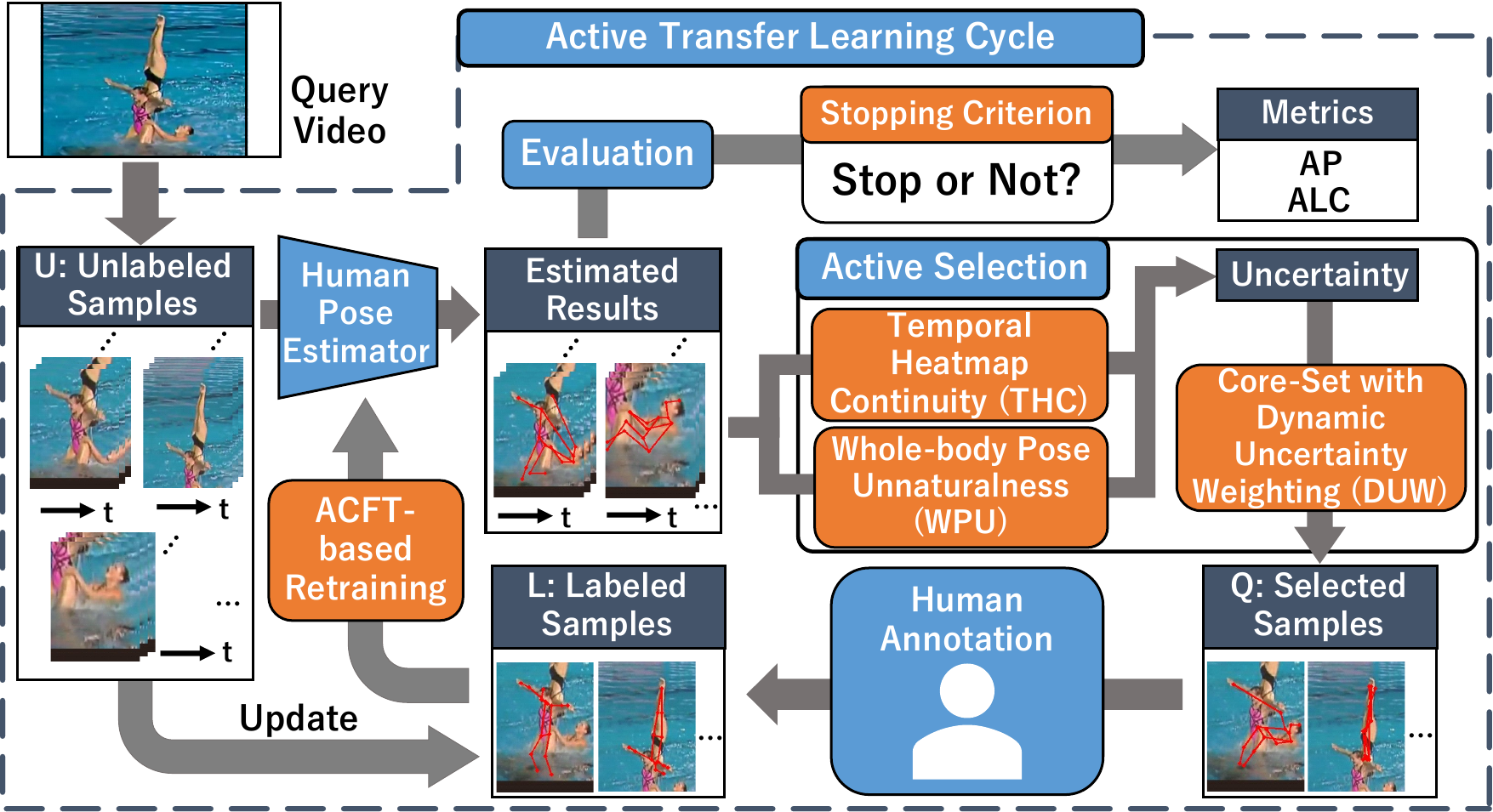}
  \end{center}
  \vspace*{-2mm}
  \captionsetup{width=\linewidth}
  \caption{
  Overview of our proposed method (Sec.~\ref{atl:overview}).
  Given an unlabeled query video, our active selection criteria (Sec.~\ref{sec:unc}) select samples to be annotated based on estimated results. Then, the HP estimator is retrained using labeled samples following ACFT~\cite{ATL_medicalimage} (Sec.~\ref{atl:acft}). The learning process terminates when the proposed SC is met (Sec.~\ref{atl:stopping}). The boxes highlighted in \textcolor{orange}{orange} represent the novelty of this study.}
  \label{fig:overview}
\end{figure*}

To this end, we propose a novel approach of applying Active Transfer Learning (ATL)~\cite{ATL_handwriting, ATL_HyperspectralImageClassification, ATL_medicalimage} on a per-video basis (see Fig.~\ref{fig:overview}). This approach involves the following two schemes, namely (i) selection of a subset of HPs for manual annotation by means of Active Learning (AL)~\cite{AL_for_HPE, ALforHPE_temporal_continuity, VL4Pose, AL4HPE_RL, AL4HPE_MATAL, ActiveImageSegmentationPropagation} and (ii) retraining of a pre-trained HP estimator with the annotated images by means of Transfer Learning (TL)~\cite{TLforHPE_rescue, TLforHPE_driver, TLforHPE_characters}.
Our contributions to ATL are as follows:
\begin{enumerate}
\item{\textbf{ACFT-based Learning (Sec.~\ref{sec:atl}):}}
    To achieve efficient ATL, we refine the existing learning approach ``Active, Continual Fine-tuning (ACFT)''~\cite{ATL_medicalimage} with a novel Stopping Criterion (SC) for video-specific ATL.

\item{\textbf{Novel Criteria (Sec.~\ref{sec:unc}):}}
    To improve the efficiency of ATL for HP estimation, we propose the following criteria for sample selection (see Fig.~\ref{fig:selected_examples}):
    \begin{itemize}
        \item{\textbf{Temporal Heatmap Continuity (THC):} 
            Uncertainty criterion based on the temporal change in the estimated heatmaps.}
        \item{\textbf{Whole-body Pose Unnaturalness (WPU):} 
            Uncertainty criterion based on the unnaturalness of the estimated full-body HP.}
    \item{\textbf{Dynamic Uncertainty Weighting (DUW):} 
            Integration of uncertainty and representativeness criteria via dynamically weighting them in Core-Set sampling~\cite{coreset} using estimated uncertainty}.
    \end{itemize}
    
\item{\textbf{Comprehensive Evaluation (Sec.~\ref{sec:exp}):}}
    We conducted quantitative and qualitative evaluations to assess our proposed ATL framework.
    Our ablation study validates that each component of our method (THC, WPU, and DUW) contributes to the performance.
    The effectiveness of our proposed SC is also demonstrated.
\end{enumerate}

\section{Related Work}
\label{sec:relwork}
\subsection{Human Pose (HP) Estimation}

Deep learning has improved HP estimation~\cite{UniPose, OTPose, PoseWarper, K-FPN, HRNet_forHPE, DBLP:journals/cviu/KawanaUHY18}. 
Simple Baseline~\cite{SimpleBaseline} and FastPose~\cite{AlphaPose} estimate 2D HPs in still images by estimating the probability distribution of each keypoint as a heatmap. For multi-person 2D HP estimation, two major approaches exist: the top-down approach~\cite{AlphaPose, DCPose, DetTrack}, where all humans in the image are first detected, then each pose is estimated; and the bottom-up approach~\cite{OpenPose, Openpifpaf}, where all keypoints in the image are first detected, then assembled into individual HPs. Our method can be applied to both single and multi-person HP estimation, regardless of whether it is top-down or bottom-up.

\subsection{Active Learning (AL)}
AL is a learning method that actively selects training data to improve the performance of a learning model, aiming to reduce annotation costs for efficiency. The selection criteria in AL are generally divided into uncertainty criteria~\cite{unc_aleatoric, HP, AL_for_HPE, ALforHPE_temporal_continuity, VL4Pose} and representativeness criteria~\cite{kmeans, coreset, ActiveImageSegmentationPropagation}. Uncertainty criteria evaluate the uncertainty of the estimation results and select samples with high uncertainty for annotation. In contrast, representativeness criteria, including Core-Set~\cite{coreset}, consider the distribution of unlabeled samples and select diverse samples. 

AL is helpful for tasks with high annotation costs, such as HP estimation~\cite{AL4HPE_RL, AL4HPE_MATAL}. Liu and Ferrari~\cite{AL_for_HPE} proposed the Multiple Peak Entropy (MPE) as a quantification of uncertainty suitable for HP estimation.
Mori~\etal~\cite {ALforHPE_temporal_continuity} proposed Temporal Pose Continuity (TPC) for HP estimation in videos, where the temporal change of the estimated pose is considered as uncertainty. We propose THC as a further extension of MPE and TPC (Sec.~\ref{unc:thc}).

Shukla~\etal~\cite{VL4Pose} modeled keypoint-level and pose-level uncertainty in the context of out-of-distribution detection. They quantified the Visual Likelihood for estimated Poses (VL4Pose) based on the skeletal model representation using a Bayesian network.  In this method, samples with lower VL4Pose are chosen as the samples with higher uncertainty.
In contrast, our proposed WPU defines uncertainty based on the anomaly scores of AutoEncoder (AE)~\cite{ICCV_AE_Anomaly, AE_anomaly, pose_anomaly}, from the perspective of anomaly detection (Sec.~\ref{unc:wpu}).

\subsection{Active Transfer Learning (ATL)}
The effectiveness of ATL has been demonstrated in tasks such as hyperspectral image classification~\cite{ATL_HyperspectralImageClassification}, and handwriting recognition~\cite{ATL_handwriting}. As an application of medical image analysis, Zhou~\etal~\cite{ATL_medicalimage} proposed Active, Continual Fine-Tuning (ACFT), an ATL method that sequentially adapts a model pre-trained on a general dataset to a different domain.
In this study, we utilize a learning method that further improves upon ACFT. Thereby, we adopt a pre-trained HP estimator from learned domains to unknown individual video domains efficiently. The procedure and our extension of ACFT are described in the next Section~\ref{sec:atl}.

\section{Video-Specific HP Estimation via ATL}
\label{sec:atl}
This section introduces our video-specific ATL framework. The overview of our method is described in Sec.~\ref{atl:overview}. In Sec.~\ref{atl:acft}, we explain our refined retraining method, an improvement upon the existing ATL approach, ACFT~\cite{ATL_medicalimage}. In Sec.~\ref{atl:stopping}, we propose a novel SC for ATL.
Details on the active selection procedure are described in Sec.~\ref{sec:unc}.

\subsection{Overview}
\label{atl:overview}
In our ATL framework, the ATL cycle (starts with $c=0$) is repeated to continuously adapt the HP estimator $M_c$ to a query video. As shown in Fig.~\ref{fig:overview}, a query video capturing human motion can be considered as a set of unlabeled samples $U = \{x_1, x_2, ..., x_N\}$, where $N$ is the total number of samples.
For example, if the number of human instances in the 1st, 2nd, and 3rd frames in the 3-frame video are 2, 4, and 1, respectively, $N = 7$.
We assume that all $N$ samples in the video are in the same domain.

The $c$-th ATL cycle starts with HP estimation on all $N$ samples within the video (``Human Pose Estimator'' in Fig.~\ref{fig:overview}). This estimation serves as the initial phase for the subsequent phases in the ATL cycle. The next phase is the ``Active Selection'' (also illustrated in Fig.~\ref{fig:overview}), in which each sample's uncertainty $C(x_i)$ where $x_i \in U$ is evaluated based on THC and WPU (described in Sec.~\ref{unc:thc} and Sec.~\ref{unc:wpu}, respectively). The computed uncertainties are then utilized in Core-Set sampling~\cite{coreset} with DUW, where uncertain and diverse samples are selected (Sec.~\ref{unc:duw}).

Following active selection, a human annotator manually annotates the chosen samples $Q$ (``Human Annotation'' in Fig.~\ref{fig:overview}). The final phase of each ATL cycle is ``ACFT-based retraining'' (Sec.~\ref{atl:acft}) of the HP estimator. In this phase, both the newly annotated samples $Q$ and already annotated samples $L$ are used for retraining $M_c$ to $M_{c+1}$. The conditions for retraining are adjusted based on the estimated results, which allows us stable continual learning.

At the end of each ATL cycle, the criterion for terminating ATL is evaluated based on the estimated results (``Stopping Criterion'' in Fig.~\ref{fig:overview}).

\subsection{ACFT-based Retraining}
\label{atl:acft}
In retraining the HP estimator, we adhere to the learning strategy proposed in ACFT~\cite{ATL_medicalimage}. That is, we continually fine-tune a pre-trained model $M_0$ with samples $L$ added via active selection. 
In the beginning, the number of labeled samples is initialized to zero (i.e., $|L| = 0$), and the HP estimator $M_0$ is pre-trained across a broad domain of a large dataset (i.e., source domain). As repeating the ATL Cycle, the HP estimator $M_c$ is adapted to the domain of each query video (i.e., target domain) continually.

However, in the original ACFT~\cite{ATL_medicalimage}, retraining is conducted over a fixed number of epochs $E$, which presents issues in terms of both performance improvement and execution time. That is, when the HP estimator is not adapted to the domain of the query video, it is necessary to increase $E$ to make drastic changes in parameters of $M_c$. Conversely, as the performance of the HP estimator improves, decreasing $E$ can reduce execution time. Following this observation, we determine the number of epochs in our ACFT-based retraining as follows:
\begin{equation}
    E_c = \alpha \times (1-G_c) \in \mathbb{N}, \label{eq:retraining}
\end{equation}
where $G_c \in [0, 1]$ represents the performance of $M_c$ for unlabeled samples $U$, and $\alpha$ is a hyperparameter.

However, in actual operation, it is impossible to correctly evaluate the estimation performance $G_c$ since we do not possess the ground truth for unlabeled samples. Therefore, an alternate metric representing the estimation performance is needed. Hence, we evaluate the performance of the estimated results only with the newly selected samples $Q$ with annotated ground truth. The generalization performance of the HP estimator $M_c$ in the $c$ th cycle is estimated as follows:
\begin{equation}
    G_c \approx \frac{1}{|Q|}{\sum_{x \in Q}OKS(x)},  \label{eq:performance}
\end{equation}
where $OKS(x) \in [0, 1]$ represents the Object Keypoint Similarity (OKS), which is an evaluation metric for HP estimation. The value of OKS increases as the estimated pose becomes more similar to the ground truth.

Additionally, we rethink the retraining of already labeled samples. The original ACFT is applied to a classification task~\cite{ATL_medicalimage}, and execution time is reduced by only retraining the misclassified labeled samples. However, since HP estimation is the regression task, it is not possible to determine misestimated data in the same way. We newly define the misestimated labeled samples $R$ as follows:
\begin{equation}
   R = \{x | OKS(x) < \theta + m \}, \label{eq:misestimated}
\end{equation}
where $\theta$ is a user-defined accuracy threshold and $m$ represents a margin set to ensure the estimation performance stably exceeds the required accuracy $\theta$.

\subsection{Stopping Criterion}
\label{atl:stopping}
In applications of AL, SC, which determines when to terminate learning, is indispensable. Typically, a common SC involves achieving the desired accuracy on cross-validation data or a hold-out set~\cite{sc_foral}, used in ACFT~\cite{ATL_medicalimage} as well. However, collecting sufficient validation data in AL is impractical because it requires much more extra annotation. SC can be also defined with agreement among multiple estimators~\cite{sc_stabilizing, sc_useradjustable} so that SC is met if the results of many estimators are equal. However, the computational cost increases in accordance with the number of estimators. While uncertainty-based SCs~\cite{sc_namedentity, sc_multi} are also proposed, the reliability of such an SC is questionable since uncertainty merely serves as an estimate of the degree of error.

Zhu and Hovy proposed the Min-error criterion~\cite{sc_zhu}, which measures the accuracy of predicted results for newly selected unlabeled samples (i.e., samples in $Q$). If the accuracy surpasses a certain threshold, the learning process is terminated. This criterion does not require additional computation and is a more reliable measure since it uses the actual error. When expressed using OKS, it is represented as follows:
\vspace{-0.1cm}
\begin{equation}  \label{eq:sc_min}
SC_{Min} =
    \begin{cases}
        1 & \frac{1}{|Q|}{\sum_{x \in Q}OKS(x)} > \theta \\
        0 & \textit{otherwise}.
    \end{cases}
\end{equation}

However, as pointed out by Zhu and Hovy~\cite{sc_zhu}, there is a risk that $SC_{Min}$ may lead to premature termination when the number of newly added unlabeled samples is small. Therefore, we propose the following $SC_{All}$, a new SC suitable for the practical use in video-specific ATL as follows:
\vspace{-0.1cm}
\begin{equation}  \label{eq:sc_all}
SC_{All} =
    \begin{cases}
        1 & \forall{x \in \{Q \cup L\}}, OKS(x) > \theta \\
        0 & \textit{otherwise}.
    \end{cases}
\end{equation}

We compare the effects of these two criteria in the experiment (Sec.~\ref{sec:exp}).

\begin{figure}[t]
  \centering
  \begin{subfigure}{\columnwidth}
    \includegraphics[width=0.85\columnwidth]{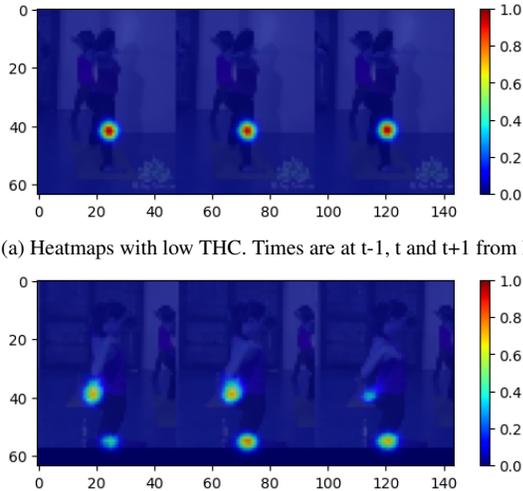}
    \caption{Heatmaps with low THC. Times are at t-1, t and t+1 from left to right.}
    \label{fig:low_thc}
  \end{subfigure}
  \begin{subfigure}{\columnwidth}
    \includegraphics[width=0.85\columnwidth]{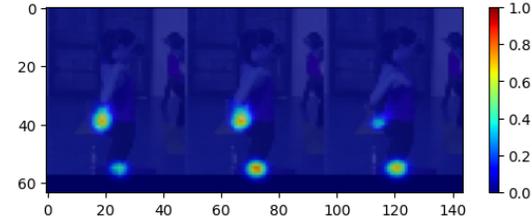}
    \caption{Heatmaps with high THC. Times are at t-1, t, and t+1 from left to right.}
    \label{fig:high_thc}
  \end{subfigure}
  \caption{Qualitative examples of our THC. (a) There is a strong peak at a single point in the heatmap between adjacent frames consistently. (b) In contrast, the estimations are inconsistent and the peaks in the heatmap are dispersed.}
  \label{fig:example_thc}
\end{figure}

\section{Uncertainty Criteria for Efficient Active Transfer Learning}
\label{sec:unc}
In the active selection phase, we actively select samples for human annotation based on the results of HP estimation for each unlabeled sample. This active selection is depicted in Fig.~\ref{fig:overview}. Our proposed uncertainty criteria, THC (Sec.~\ref{unc:thc}) and WPU (Sec.~\ref{unc:wpu}) calculate the uncertainty from the heatmaps obtained during HP estimation and the estimated whole-body poses, respectively. These uncertainties serve as weights in our DUW Core-Set~\cite{coreset} approach (Sec.~\ref{unc:duw}).
This approach guides the selection of diverse and uncertain samples effectively.

\subsection{THC: Uncertainty Criterion based on Temporal Heatmap Continuity}
\label{unc:thc}
In this section, we introduce a new uncertainty criterion called THC, an extension of MPE~\cite{AL_for_HPE} and TPC~\cite{ALforHPE_temporal_continuity}.

Many HP estimation methods output heatmaps for each keypoint and select the maximum probability position in the heatmap as the keypoint position. The rest of the information is discarded although it contains valuable information for assessing the estimation results. For instance, a single peak with low variance might indicate a confident estimation. In contrast, the presence of multiple peaks with high variance could be a signal of an uncertain estimation.

Based on this concept, MPE~\cite{AL_for_HPE} was proposed as an uncertainty criterion to replace conventional methods such as Least Confidence (LC)~\cite{HP}, which quantify uncertainty only from the maximum values in the heatmap. MPE utilizes not only the maximum values but also local peaks in the heatmap to calculate entropy, making use of the rich information of the heatmap.

Furthermore, when evaluating HP estimation results in videos, we can utilize the property that `the correct keypoint position does not change significantly between adjacent frames'. TPC~\cite{ALforHPE_temporal_continuity} was proposed based on this idea, quantifying the uncertainty of HP estimation in videos by the temporal change of the estimated poses. It sums up the Euclidean distances of estimated keypoint positions between adjacent frames, considering larger values to indicate higher uncertainty.

We introduce a new uncertainty criterion extending upon both MPE and TPC, namely, THC. This quantifies uncertainty by utilizing spatially rich information in heatmaps. In addition, THC considers the continuity in estimation results between temporally adjacent frames.

Specifically, the uncertainty is quantified by calculating the Sum of the Absolute Difference (SAD) of estimated heatmaps for each keypoint between adjacent frames, as shown below:
\vspace{-0.1cm}
\begin{align}
    C_{THC}(\it{F_t}) &=  \frac{1}{K}{\sum_{k=1}^K}{SAD({H^k_{t-1}},H^k_t)}  \nonumber \\
    & + \frac{1}{K}{\sum_{k=1}^K}{SAD(H^k_t,{H^k_{t+1}})} \label{eq:THC} \\
    &=  \frac{1}{K}{\sum_{k=1}^K}{\sum_{{p^k_t}\in{H^k_t}}}(|{p^k_{t-1}} - p^k_t|  \nonumber \\
    & + |p^k_t - {p^k_{t+1}}|),  \label{eq:sad} 
\end{align}
where $K$, $H^k_t$, and $p^k_t$ denote the number of keypoints in each HP, the estimated heatmap for keypoint $k$ in frame $t$, and the probability at each position in $H^k_t$, respectively. Examples of heatmaps with low THC and high THC are shown in Fig.~\ref{fig:example_thc}.

\begin{figure}[t]
  \centering
  \begin{center}
    \includegraphics[clip, width=\columnwidth]{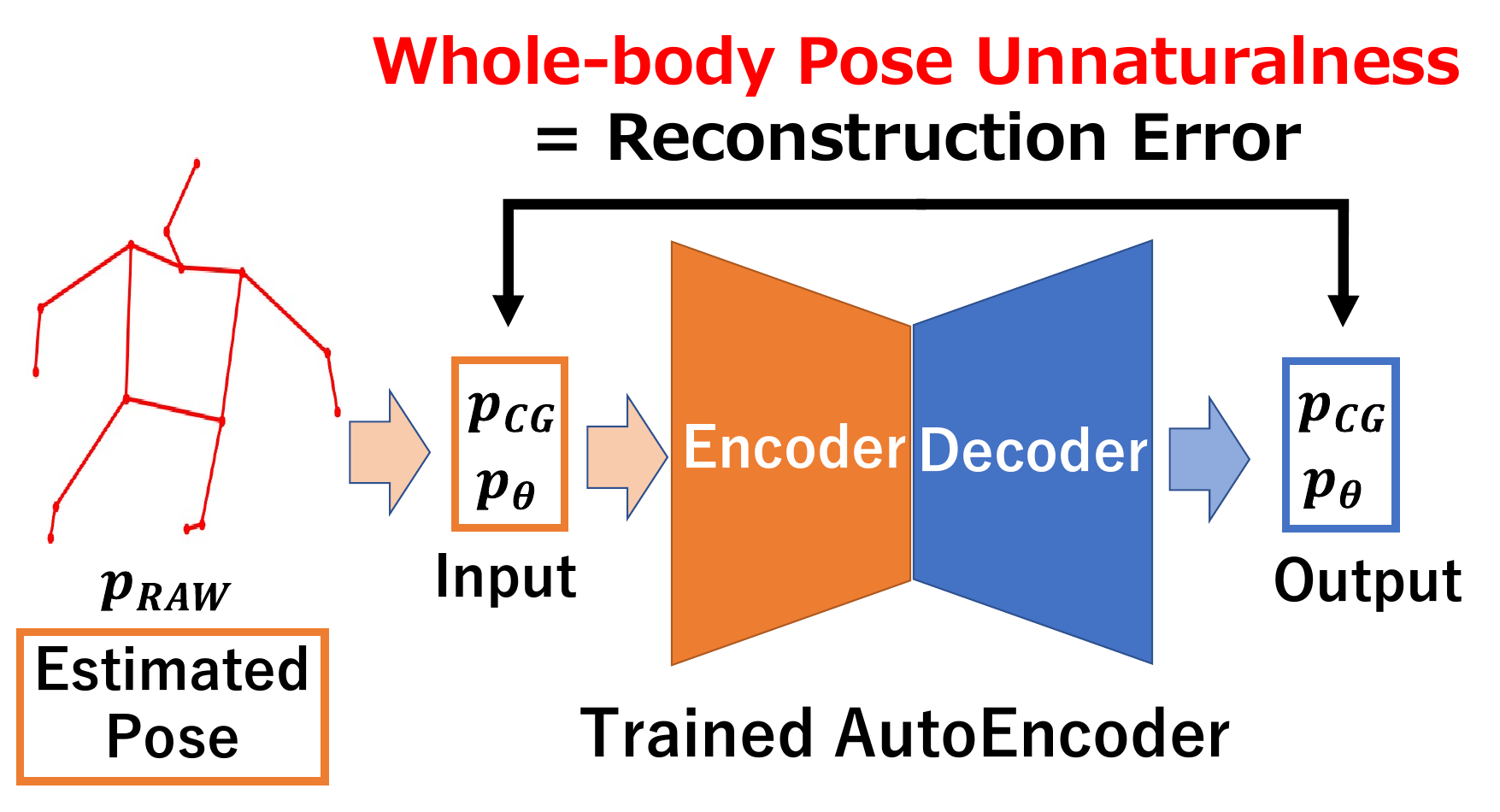}
  \end{center}
  \caption{Calculation of Whole-body Pose Unnaturalness (WPU). Given the estimated HP, its hybrid feature~\cite{Hybrid_feature} is calculated and fed into the trained AutoEncoder (AE). WPU is defined by the reconstruction error between the input and output of AE.}
  \label{fig:WPU}
\end{figure}

\subsection{WPU: Uncertainty Criterion based on Whole-body Pose Unnaturalness}
\label{unc:wpu}
Previous AL for HP estimation~\cite{AL_for_HPE, ALforHPE_temporal_continuity, DBLP:journals/cviu/UkitaU18} computes the uncertainty of an entire body pose by summing the uncertainty of each keypoint. However, this approach may miss unnatural poses where individual keypoints exhibit low uncertainty. Therefore, we introduce WPU, a novel uncertainty criterion that quantifies the unnaturalness of whole-body poses in the context of anomaly detection.

In our approach, we train AE~\cite{ICCV_AE_Anomaly, AE_anomaly, pose_anomaly}, a simple anomaly detection model, on ground-truth HPs from the dataset. The trained AE successfully reconstructs natural HPs but struggles to reconstruct unnatural ones, resulting in larger reconstruction errors. Therefore, this error can measure pose-level uncertainty, as shown in Fig.~\ref{fig:WPU}. 

Although VL4Pose~\cite{VL4Pose} also tackled the issue of pose-level uncertainty using a Bayesian Neural Network architecture, our WPU has a significant advantage in model efficiency. While VL4Pose utilizes $\approx$ 2.1M parameters, WPU only needs 2.6K parameters. This results in a simple but effective uncertainty measure with fewer computational costs.

To train the AE well, we do not use raw keypoint coordinates from the dataset but rather calculate the Hybrid feature~\cite{Hybrid_feature} for inputs. This Hybrid feature is a pose representation robust against scaling differences and rotations, consisting of the Center of Gravity (CG) feature $\it{p_{CG}}$\cite{Hybrid_CenterOfGravity} and the Angle feature $\it{p_\theta}$\cite{Hybrid_Angle}, which represents eight critical joint angles. 

Furthermore, through ATL procedure, the pre-trained AE is retrained on labeled poses in the query video. By learning natural poses in the target domain (i.e., the query video) during ATL, the AE can learn the feature of natural poses that are not included in the source domain (i.e., the training dataset).

\begin{figure}[t]
    \centering
    \includegraphics[width=\columnwidth]{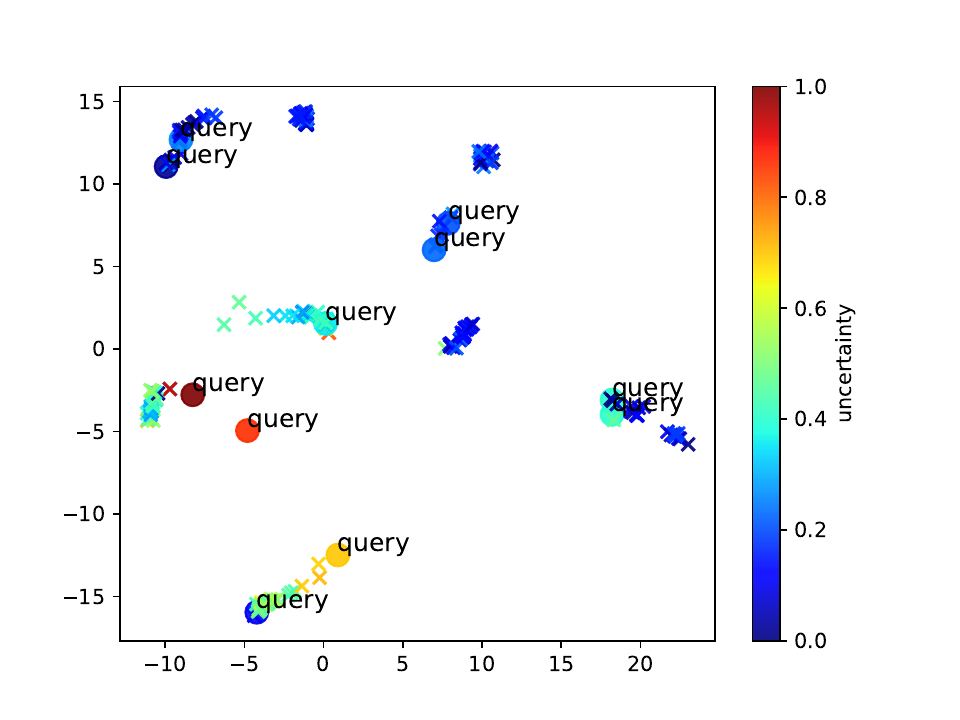}
    \caption{A visualization result of the sample selection of our proposed criterion (THC+WPU+DUW). We have utilized DensMAP~\cite{densmap} to plot feature vectors extracted by the HP estimator. In this plot, circles represent newly selected samples, while cross marks denote unlabeled samples that were not selected for the current ATL cycle. The color of the plot corresponds to the normalized uncertainty.}
    \label{fig:example_duw}
\end{figure}

\subsection{DUW: Dynamic Uncertainty Weighting of Core-Set sampling}
\label{unc:duw}
Many AL studies~\cite{VL4Pose, AL_for_HPE, ALforHPE_temporal_continuity} tend to rely solely on uncertainty for sample selection, but it can fall into selection bias~\cite{al_bias, ALSurvey}. This is largely because uncertainty criteria tend to select similar data, like redundant samples from continuous video frames in our setting.

Therefore, in this research, we propose DUW, which balances both uncertainty and diversity by extending the acquisition function of Core-Set sampling~\cite{coreset}. In the original Core-Set sampling, $u$, a new sample to be labeled is sequentially selected according to the following acquisition function:
\begin{equation}
     u = \argmax_{i \in U} \min_{j \in L} \Delta(x_i, x_j), \label{eq:coreset}
\end{equation}
where $\Delta(x_i, x_j)$ is the Euclidean distance between sample's feature vectors $x_i$ and $x_j$.

In contrast, we define the acquisition function in DUW based on each sample's uncertainty score $C(x_i)$ in the following way:
\begin{equation}
     u = \argmax_{i \in U} \{\min_{j \in L} \{(1-G_c) \times \Delta(x_i, x_j)\} + G_c \times \lambda C(x_i)\}, \label{eq:duw}
\end{equation}

Here, $G_c$ is approximated by Eq.~(\ref{eq:performance}) and $\lambda$ is a hyperparameter, respectively.

As depicted in Eq.~(\ref{eq:duw}), when $\lambda$ is 0, the sample selection is equal to the original Core-Set sampling~\cite{kmeans}. Conversely, when $\lambda$ is large, sample selection is heavily influenced by the uncertainty score. The balance between uncertainty and diversity dynamically changes based on $G_c$ too. When $G_c$ is low, the selection prioritizes coverage over the whole samples, while it emphasizes uncertain estimations when $G_c$ is high. Since $G_c$ increases through ATL, representative samples tend to be selected in the initial phase, and uncertain samples come in selection as ATL progresses. With this formulation, we aim to rapidly cover the data distribution within the query video at the initial cycles of ATL and subsequently promote the identification of remaining hard samples through uncertainty measurement. The example of informative and diverse sample selection using DUW is shown in Fig.~\ref{fig:example_duw}.

\begin{table}[t]
  \caption{Quantitative results of our proposed video-specific ATL on PoseTrack21~\cite{PoseTrack21}. \red{Red} and \blue{blue} indicate the best and the second best, respectively. AP@0.6 is the average AP of 170 test videos with a 0.6 OKS threshold. ``5\%'' means the estimation result with 5\% labeled samples in the query video. ALC values are also calculated by an average of 170 test videos.}
  \centering
    \begin{center}
        \begin{tabular}{l| c c c | c}
        \hline
        \multicolumn{1}{l|}{\multirow{2}{*}{Criterion}} & \multicolumn{3}{c}{\makebox[12mm]{AP@0.6 (\%)}} & \multicolumn{1}{|c}{ALC} \\
        \multicolumn{1}{l|}{} & 5\% & 20\% & 40\% & \multicolumn{1}{|c}{(\%)} \\
        \hline
        Random & 87.76 & 96.09 & 97.39 & 96.91 \\
        LC~\cite{HP} & 77.49 & 94.60 & 96.77 & 95.74 \\
        MPE~\cite{AL_for_HPE} & 78.96 & 95.09 & 97.23 & 96.11 \\
        TPC~\cite{ALforHPE_temporal_continuity} & 83.38 & 95.32 & 97.31 & 96.40 \\
        k-means~\cite{kmeans} & \textbf{\red{93.97}} & 96.37 & 98.11 & 97.65 \\
        Core-Set~\cite{coreset} & 93.18 & \textbf{\blue{97.62}} & \textbf{\blue{98.60}} & \textbf{\blue{98.12}} \\
        \begin{tabular}[c]{@{}l@{}}\textbf{Ours}\\ \textbf{(THC+WPU+DUW)}\end{tabular} & \textbf{\blue{93.35}} & \textbf{\red{97.90}} & \textbf{\red{98.77}} & \textbf{\red{98.21}} \\
        \hline
        \end{tabular}
    \end{center}
    \label{table:PoseTrack21}
    \vspace{-0.2cm}
\end{table}

\section{Experiments}
\label{sec:exp}
This section is broken down into five parts: (1) We outline evaluation metrics and specify our implementation (Sec.~\ref{exp:evalandimpl}). (2) We introduce the various selection criteria used, including our proposed method (Sec.~\ref{exp:criteria}). (3) We contrast our approach with the baseline and several state-of-the-art methods (Sec.~\ref{exp:sota}). (4) We conduct an ablation study to verify the effect of each component in our framework (Sec.~\ref{exp:ablation}).
(5) Lastly, we examine the effectiveness of our newly proposed SC (Sec.~\ref{exp:sc}).

\begin{figure}[t]
  \centering
  \vspace{-0.3cm}
  \begin{center}
    \includegraphics[clip, width=\columnwidth]{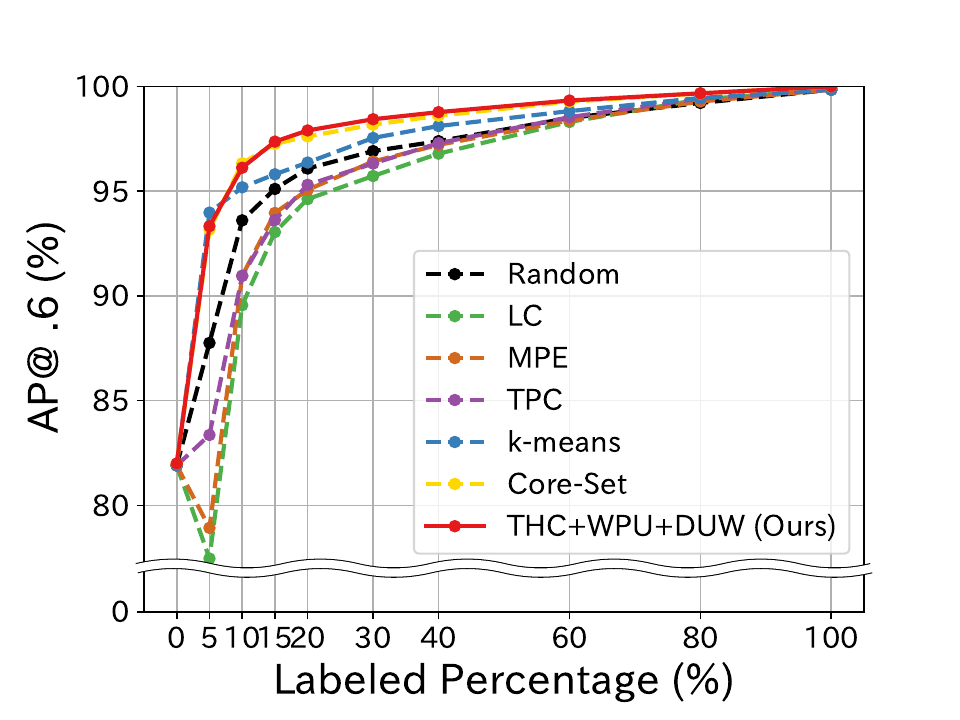}
  \end{center}
  \caption{Learning Curve of video-specific ATL on PoseTrack21~\cite{PoseTrack21}. }
  \label{fig:PoseTrack21result}
  \vspace{-0.15cm}
\end{figure}

\subsection{Evaluation and Implementation Details}
\label{exp:evalandimpl}
\textbf{Dataset.}
Two large-scale datasets for multi-person 2D HP estimation, PoseTrack21~\cite{PoseTrack21} and JRDB-Pose~\cite{JRDB-Pose} were employed for our experiments.

PoseTrack21~\cite{PoseTrack21} consists of 593 training videos and 170 validation videos. Since PoseTrack21 does not provide its test data labels, we utilized 579 out of 593 original training videos for training and the remaining 14 videos for validation and conducted evaluations using the original 170 validation videos. For each video evaluation, about 30 annotated frames were utilized. A skeleton-based pose representation in PoseTrack21 consists of 15 keypoints.

In terms of JRDB-Pose~\cite{JRDB-Pose}, as well as PoseTrack21, it does not have available labels for the test split. Thus, we utilized the 27 provided videos, dividing them into 10 for training, 2 for validation, and 15 for testing. For each video, we used the first 150 frames (five times the number for PoseTrack21~\cite{PoseTrack21}) extracted from stitched images. Poses in JRDB-Pose consist of 17 keypoints.

\textbf{HP Estimator.}
In the video-specific ATL, we followed the top-down approach manner~\cite{AlphaPose, DCPose, DetTrack}. To simplify the process, we used ground truth bounding boxes and tracking IDs as detection results. Simple Baseline~\cite{SimpleBaseline} and FastPose~\cite{AlphaPose} were employed to estimate HP in PoseTrack21~\cite{PoseTrack21} and JRDB-Pose~\cite{JRDB-Pose}, respectively. We pre-trained both models on the 579 training videos from PoseTrack21~\cite{PoseTrack21} using the Adam optimizer~\cite{Adam} with a learning rate of $1.0 e^{-3}$ and 62/57 epochs terminated by early-stopping, respectively. Training data are augmented by flipping, rotation, and scaling. Subsequently, we fine-tuned this pre-trained FastPose on 10 videos from JRDB-Pose with $lr = 5.0 e^{-4}$ and 40 epochs.

\textbf{Evaluation Metrics.}
The efficacy of our proposed video-specific ATL was evaluated using Average Precision (AP) and Area under the Learning Curve (ALC)~\cite{Area_Under_the_Learning_Curve, ATL_medicalimage}. The AP was used to assess the HP estimation results at each ATL cycle. We adopted the MS COCO's calculation~\cite{MSCOCO}, using OKS to determine the accuracy of the estimated pose. The ALC was used to evaluate the overall ATL efficiency. The ALC is calculated from a graph, where the vertical axis represents AP (\%) and the horizontal axis represents the percentage of labeled samples (\%). It should be noted that in the calculation of AP and ALC, poses annotated during ATL are considered correctly estimated. That is to say, AP surely reaches 100\% when 100\% HPs are labeled.

\textbf{Human Annotation.}
Consistent with other AL studies~\cite{AL_for_HPE, ALforHPE_temporal_continuity}, we simulated human annotation by automatically providing ground truth annotations for selected samples. The annotated samples are incrementally added for retraining, as illustrated in Fig.~\ref{fig:overview}. The [5\%, 5\%, 5\%, 5\%, 10\%, 10\%, 20\%, 20\%, 20\%] of the total HPs in the query video are sequentially added at each ATL cycle.

\textbf{Retraining.}
For retraining the HP estimator, we used the AdamW~\cite{AdamW} optimizer with learning rate $= 2.5 e^{-4}$, weight decay $= 0.7$, and $\gamma = 0.99$, respectively. $\alpha$ in Eq.~(\ref{eq:retraining}) was set to 250. To prevent overfitting, data augmentation techniques such as flipping, rotation, and scaling were applied to the retrained samples. 

\begin{table}[t]
  \caption{Ablation study results of video-specific ATL on PoseTrack21~\cite{PoseTrack21}. \red{Red} and \blue{blue} indicate the best and the second best, respectively. AP@0.6 is the average AP of 170 test videos with a 0.6 OKS threshold. ``5\%'' means the estimation result with 5\% labeled samples. ALC is also the average of 170 test videos.}
  \centering
  \begin{center}
    \begin{tabular}{l| r r r | r}
    \hline
    \multicolumn{1}{l|}{\multirow{2}{*}{Criterion}} & \multicolumn{3}{c}{\makebox[12mm]{AP@0.6 (\%)}} & \multicolumn{1}{|c}{\makebox[8mm]{ALC}} \\
    \multicolumn{1}{l|}{} & 5\% & 20\% & 40\% & \multicolumn{1}{|c}{(\%)} \\
    \hline
    Core-Set~\cite{coreset} & 93.18 & 97.62 & 98.60 & 98.12 \\
    \hline
    THC & 82.59 & 92.86 & 96.43 & 95.45 \\
    WPU & 85.56 & 94.74 & 97.31 & 96.45 \\
    THC+WPU & 84.82 & 95.17 & 97.25 & 96.51 \\
    THC+DUW & 93.12 & 97.70 & \textbf{\blue{98.91}} & 98.19 \\
    WPU+DUW & \textbf{\blue{93.19}} & \textbf{\blue{97.87}} & 98.76 & 98.17 \\
    \hline
    THC+WPU+DUW & \multirow{2}{*}{93.02} & \multirow{2}{*}{97.68} & \multirow{2}{*}{98.81} & \multirow{2}{*}{98.14} \\
    (fixed) &  & & & \\
    \hline
    THC+WPU+DUW & \multirow{2}{*}{93.18} & \multirow{2}{*}{97.86} & \multirow{2}{*}{98.80} & \multirow{2}{*}{98.16} \\
    (increase) &  & & & \\
    THC+WPU+DUW & \multirow{2}{*}{\textbf{\red{93.35}}} & \multirow{2}{*}{\textbf{\red{97.90}}} & \multirow{2}{*}{98.77} & \multirow{2}{*}{\textbf{\blue{98.21}}} \\
    (const) &  & & & \\
    THC+WPU+DUW & \multirow{2}{*}{93.08} & \multirow{2}{*}{97.72} & \multirow{2}{*}{\textbf{\red{98.94}}} & \multirow{2}{*}{\textbf{\red{98.24}}} \\
    (decrease) &  & & & \\
    \hline
    \end{tabular}
  \end{center}
    \label{table:ablation}
    \vspace{-0.6cm}
\end{table}

\subsection{Active Selection Criteria}
\label{exp:criteria}
For comparative experiments, our proposed method and several ATL approaches are implemented with the following selection criteria:
\begin{itemize}[noitemsep, topsep=1pt, parsep=1pt, partopsep=1pt]
    \item \textbf{Random:} Random uniform sampling.
    \item \textbf{LC:} A traditional uncertainty measurement described in~\cite{HP}. The implementation followed~\cite{AL_for_HPE}.
    \item \textbf{MPE:} An uncertainty criterion in~\cite{AL_for_HPE}.
    \item \textbf{TPC:} An uncertainty criterion in~\cite{ALforHPE_temporal_continuity}.
    \item \textbf{k-means:} A representativeness criterion used in~\cite{kmeans}.
    \item \textbf{Core-Set:} An original Core-Set sampling in~\cite{coreset}. This implementation followed~\cite{DeepAL}.
    \item \textbf{THC:} Our uncertainty criterion proposed in Sec.~\ref{unc:thc}.
    \item \textbf{WPU:} Our uncertainty criterion proposed in Sec.~\ref{unc:wpu}. For the experiment on PoseTrack21~\cite{PoseTrack21}, the AE was trained with the ground truth keypoints from the 579 training videos (300 epochs with learning rate $= 1.0 e^{-3}$ by Adam~\cite{Adam}). For JRDB-Pose~\cite{JRDB-Pose}, pre-training was conducted using 10/2 videos in the train/val set. Both the encoder and the decoder of the AE have four layers each, and the dimension of the latent variables is 4. The AE is retrained at each ATL cycle by Adam (20 epochs with $lr = 8.0 e^{-4}$).
    \item \textbf{DUW:} The combination of uncertainty criteria and Core-Set sampling~\cite{coreset} proposed in Sec.~\ref{unc:duw}. The value of $\lambda$ in Eq.~(\ref{eq:duw}) was set to 0.01 and 1000 for PoseTrack21~\cite{PoseTrack21} and JRDB-Pose based on a hyperparameter search based on the performance of video-specific ATL on validation videos, respectively. We set the weights of THC and WPU at a 1:1 ratio in Sec.~\ref{exp:sota}.
    
\end{itemize}

For methods that perform sample selection based solely on uncertainty, samples with higher $C(x_i)$ are prioritized to be added to the labeled data. For further details, please refer to our codebase at: \url{https://github.com/ImIntheMiddle/VATL4Pose-WACV2024}

\subsection{Baseline and State-of-the-art Comparison}
\label{exp:sota}
Figure~\ref{fig:PoseTrack21result} and Table~\ref{table:PoseTrack21} show quantitative results of the proposed video-specific ATL on PoseTrack21~\cite{PoseTrack21}. While the performances of all uncertainty-based methods~\cite{HP, AL_for_HPE, ALforHPE_temporal_continuity} are less than the random selection, our method (``THC+WPU+DUW'') outperforms other methods throughout the entire ATL process. Video-specific ATL with our methods can efficiently achieve accurate HP estimation (e.g., as shown in Table~\ref{table:PoseTrack21}, our method got AP@0.6 $\approx$ 98\% with only 20\% labeled samples).

The results for the 15 test videos from JRDB-Pose are presented in Table~\ref{table:JRDBPose_main}. Here too, our proposed method (``Ours'') achieved performance close to 90\% with only 5\% of the labeling. Furthermore, the ALC performance of the proposed method stably outperforms comparative methods across the entire ATL procedure.
For the complete table and evaluation with another metric, please refer to Sec.~B in Appendix.

\begin{table}[t]
  \caption{Quantitative results of our proposed video-specific ATL on JRDB-Pose~\cite{JRDB-Pose}. \red{Red} and \blue{blue} indicate the best and the second best, respectively. AP is an average of 15 test videos. ``5\%'' means the estimation result with 5\% labeled samples in the query video. ALC values are also an average of 15 test videos.}
  \centering
    \begin{center}
        \begin{tabular}{l| c c c | c}
        \hline
        \multicolumn{1}{l|}{\multirow{2}{*}{Criterion}} & \multicolumn{3}{c}{\makebox[12mm]{AP@0.6 (\%)}} & \multicolumn{1}{|c}{ALC} \\
        \multicolumn{1}{l|}{} & 5\% & 20\% & 40\% & \multicolumn{1}{|c}{(\%)} \\
        \hline
        Random & 88.16 & 94.19 & 96.46 & 95.42 \\
        LC~\cite{HP} & 65.04 & 89.34 & 94.84 & 92.67 \\
        MPE~\cite{AL_for_HPE} & 81.78 & 95.74 & \textbf{\red{98.03}} & 95.76 \\
        TPC~\cite{ALforHPE_temporal_continuity} & 74.83 & 92.25 & 95.74 & 93.76 \\
        k-means~\cite{kmeans} & \textbf{\blue{88.97}} & \textbf{\blue{95.98}} & 97.53 & \textbf{\blue{96.41}} \\
        Core-Set~\cite{coreset} & 85.09 & 95.27 & 96.80 & 95.60 \\
        \begin{tabular}[c]{@{}l@{}}\textbf{Ours}\\ \textbf{(THC+WPU+DUW)}\end{tabular} & \textbf{\red{89.76}} & \textbf{\red{96.48}} & \textbf{\blue{97.59}} & \textbf{\red{96.52}} \\
        \hline
        \end{tabular}
    \end{center}
    \label{table:JRDBPose_main}
    \vspace{-0.4cm}
\end{table}

\subsection{Ablation Studies}
\label{exp:ablation}
Table~\ref{table:ablation} shows the ablation study results of video-specific ATL on PoseTrack21~\cite{PoseTrack21}. Our proposed methods, THC, WPU, and DUW are all used together, resulting in the highest ALC. THC+DUW and WPU+DUW surpassed the performance of the original Core-Set~\cite{coreset} due to the incorporation of uncertainty in sample selection.

When comparing only uncertainty, THC+WPU achieves the highest ALC including the other methods in Table~\ref{table:PoseTrack21}. This suggests the effect of combining THC with WPU.

Next, we compared our method with a case where the balance between uncertainty and representativeness in Eq.~(\ref{eq:duw}) is not dynamically adjusted by $G_c$, only using a fixed hyperparameter, $\lambda$. $\lambda$ was 1 based on a hyperparameter search using video-specific ATL on the 14 validation videos. While the results (THC+WPU+DUW (fixed)) surpassed Core-Set~\cite{coreset}, it is poorer than other results of THC+WPU+DUW, using the dynamic weighting by $G_c$. This demonstrates the effectiveness of dynamically adjusting the balance between uncertainty and representativeness.

Lastly, we conducted a detailed ablation study on the combination of THC and WPU. In this study, we compared three scenarios: increasing the proportion of THC linearly from 0 to 1 (``increase'') based on the labeled percentage, setting the same weight for THC and WPU (``const''), and decreasing the proportion of THC linearly from 1 to 0 (``decrease''). As shown in Tab.~\ref{table:ablation}, it is evident that setting the balance between THC and WPU to ``const'' enjoys significant performance improvement during the early stages and the midpoint of ATL. Nevertheless, the performance of ``decrease'' tends to be enhanced in the later stages. These findings further motivate the need to design appropriate strategies for combining THC and WPU.

For further detailed analysis and results, please refer to Sec.~A, C and D in Appendix.

\subsection{Effectiveness of proposed SC}
\label{exp:sc}
Table~\ref{table:sc} compares existing SC, Min-error~\cite{sc_zhu}, with our $SC_{All}$ proposed in Sec.~\ref{atl:stopping}. The labeled samples were increased by 10\% increments from 0\% and the margin $m$ in Eq.~(\ref{eq:misestimated}) was set to 0.05. As hypothesized in Sec.~\ref{atl:stopping}, Min-error relies on the average value of a small number of samples and encounters premature stops at $<$ 97\% AP. On the other hand, our $SC_{All}$ terminates ATL when AP has almost reached 100\% for any threshold, thereby ensuring the HP estimation accuracy that the user demands more precisely.

\begin{table}[t]
    \caption{Effectiveness of our SC. $\theta$ is the target value of OKS defined by the user. AP@$\theta$ represents the value of AP calculated at the time learning stopped, with $\theta$ as the threshold. ``Stopped'' and ``Actual'' denote the labeled percentage when learning was stopped by the SC and the labeled percentage when all samples actually reached an OKS above $\theta$, respectively.}
    \centering
    \begin{center}
        \begin{tabular}{l|c|c|c|c}
        \hline
        SC & $\theta$ & AP@$\theta$ (\%)& Stopped (\%) & Actual (\%)\\
        \hline
        \multirow{4}{*}{\begin{tabular}[l]{@{}l@{}}$SC_{Min}$\\ \cite{sc_zhu}\end{tabular}}
         & 0.5 & \textbf{\blue{96.86}} & 10.04 & 36.90 \\
         & 0.6 & \textbf{\blue{96.19}} & 10.85 & 40.26 \\
         & 0.7 & \textbf{\blue{95.75}} & 12.55 & 46.55 \\
         & 0.8 & \textbf{\blue{95.58}} & 16.94 & 56.54 \\
        \hline
        \multirow{4}{*}{\begin{tabular}[l]{@{}l@{}}$SC_{All}$\\ (Ours)\end{tabular}}
         & 0.5 & \textbf{\red{99.25}} & 29.61 & 36.90 \\
         & 0.6 & \textbf{\red{99.55}} & 33.38 & 40.26 \\
         & 0.7 & \textbf{\red{99.60}} & 39.61 & 46.55 \\
         & 0.8 & \textbf{\red{99.50}} & 49.46 & 56.54 \\
        \hline
        \end{tabular}
    \end{center}
    \vspace{-0.4cm}
    \label{table:sc}
\end{table}

\section{Concluding Remarks}
\label{sec:conc}
In this study, we addressed video-specific HP estimation using ATL for the first time. We revisited the existing ATL method~\cite{ATL_medicalimage} and proposed a retraining method suitable for video-specific ATL along with novel SC. To enhance learning efficiency, we proposed three novel selection criteria: THC, WPU, and DUW.

Our criteria outperformed existing methods by enabling the selection of uncertain and diverse samples.
Additionally, we found that proposed $SC_{All}$ can accurately determine the timing to terminate ATL for practical use.

For future work, we suggest integrating video-based HP estimation methods to enhance performance~\cite{PoseWarper, K-FPN, OTPose, DeciWatch}. Additionally, utilizing semi-supervised learning~\cite{SemSeg_with_ActiveSemi-SupervisedLearning, feng2023rethinking, Seq-UPS, DBLP:journals/cviu/UkitaU18} could further reduce annotation cost.

\clearpage

{\small
\bibliographystyle{ieee_fullname}
\bibliography{ref}

\begin{thebibliography}{10}\itemsep=-1pt

\bibitem{HPE_in_succer_players}
Reza Afrouzian, Hadi Seyedarabi, and Shohreh Kasaei.
\newblock Correction to: Pose estimation of soccer players using multiple uncalibrated cameras.
\newblock {\em Multim. Tools Appl.}, 78(2):2641, 2019.

\bibitem{UniPose}
Bruno Artacho and Andreas~E. Savakis.
\newblock Unipose: Unified human pose estimation in single images and videos.
\newblock In {\em CVPR}, 2020.

\bibitem{pose_cvsports}
Tobias Baumgartner and Stefanie Klatt.
\newblock Monocular 3d human pose estimation for sports broadcasts using partial sports field registration.
\newblock In {\em CVSports (CVPRW)}, 2023.

\bibitem{PoseWarper}
Gedas Bertasius, Christoph Feichtenhofer, Du Tran, Jianbo Shi, and Lorenzo Torresani.
\newblock Learning temporal pose estimation from sparsely-labeled videos.
\newblock In {\em NeurIPS}, 2019.

\bibitem{sc_stabilizing}
Michael Bloodgood and John Grothendieck.
\newblock Analysis of stopping active learning based on stabilizing predictions.
\newblock In {\em CoNLL}, 2013.

\bibitem{sc_useradjustable}
Michael Bloodgood and K. Vijay{-}Shanker.
\newblock A method for stopping active learning based on stabilizing predictions and the need for user-adjustable stopping.
\newblock In {\em CoNLL}, 2009.

\bibitem{ATL_handwriting}
Eric Burdett, Stanley Fujimoto, Timothy Brown, Ammon Shurtz, Daniel Segrera, Lawry Sorenson, Mark~J. Clement, and Joseph Price.
\newblock Active transfer learning for handwriting recognition.
\newblock In {\em ICFHR}, 2022.

\bibitem{OpenPose}
Zhe Cao, Tomas Simon, Shih{-}En Wei, and Yaser Sheikh.
\newblock Realtime multi-person 2d pose estimation using part affinity fields.
\newblock In {\em CVPR}, 2017.

\bibitem{ASD2}
Claire Chambers, Nidhi Seethapathi, Rachit Saluja, Helen Loeb, Samuel Pierce, Daniel Bogen, Laura Prosser, Michelle Johnson, and Konrad Kording.
\newblock Computer vision to automatically assess infant neuromotor risk.
\newblock {\em {IEEE} Trans. Neural Syst. Rehabil. Eng.}, 28:2431--2442, 2020.

\bibitem{HPE_in_healthcare}
Kenny Chen, Paolo Gabriel, Abdulwahab Alasfour, Chenghao Gong, Werner~K. Doyle, Orrin Devinsky, Daniel Friedman, Patricia Dugan, Lucia Melloni, Thomas Thesen, David Gonda, Shifteh Sattar, Sonya Wang, and Vikash Gilja.
\newblock Patient-specific pose estimation in clinical environments.
\newblock {\em {IEEE} J. Transl. Eng. Health Med.}, pages 1--11, 2018.

\bibitem{TLforHPE_characters}
Shuhong Chen and Matthias Zwicker.
\newblock Transfer learning for pose estimation of illustrated characters.
\newblock In {\em WACV}, 2022.

\bibitem{AE_anomaly}
Zhaomin Chen, Chai~Kiat Yeo, Bu{-}Sung Lee, and Chiew~Tong Lau.
\newblock Autoencoder-based network anomaly detection.
\newblock In {\em WTS}, 2018.

\bibitem{HPE_in_Surveillance}
Mickael Cormier, Aris Clepe, Andreas Specker, and Jürgen Beyerer.
\newblock Where are we with human pose estimation in real-world surveillance?
\newblock In {\em RWS (WACVW)}, 2022.

\bibitem{pose_surveillance}
Mickael Cormier, Fabian R{\"{o}}pke, Thomas Golda, and J{\"{u}}rgen Beyerer.
\newblock Interactive labeling for human pose estimation in surveillance videos.
\newblock In {\em ILDAV (ICCVW)}, 2021.

\bibitem{Area_Under_the_Learning_Curve}
Matt Culver, Kun Deng, and Stephen Scott.
\newblock Active learning to maximize area under the {ROC} curve.
\newblock In {\em ICDM}, 2006.

\bibitem{ATL_HyperspectralImageClassification}
Cheng Deng, Yumeng Xue, Xianglong Liu, Chao Li, and Dacheng Tao.
\newblock Active transfer learning network: {A} unified deep joint spectral-spatial feature learning model for hyperspectral image classification.
\newblock {\em {IEEE} Trans. Geosci. Remote. Sens.}, 57(3):1741--1754, 2019.

\bibitem{PoseTrack21}
Andreas Doering, Di Chen, Shanshan Zhang, Bernt Schiele, and Juergen Gall.
\newblock Posetrack21: {A} dataset for person search, multi-object tracking and multi-person pose tracking.
\newblock In {\em CVPR}, 2022.

\bibitem{TLforHPE_rescue}
Carl Doersch and Andrew Zisserman.
\newblock Sim2real transfer learning for 3d human pose estimation: motion to the rescue.
\newblock In Hanna~M. Wallach, Hugo Larochelle, Alina Beygelzimer, Florence d'Alch{\'{e}}{-}Buc, Emily~B. Fox, and Roman Garnett, editors, {\em NeurIPS}, 2019.

\bibitem{AlphaPose}
Hao-Shu Fang, Jiefeng Li, Hongyang Tang, Chao Xu, Haoyi Zhu, Yuliang Xiu, Yong-Lu Li, and Cewu Lu.
\newblock Alphapose: Whole-body regional multi-person pose estimation and tracking in real-time.
\newblock {\em {IEEE} Trans. Pattern Anal. Mach. Intell.}, 2022.

\bibitem{al_bias}
Sebastian Farquhar, Yarin Gal, and Tom Rainforth.
\newblock On statistical bias in active learning: How and when to fix it.
\newblock In {\em ICLR}, 2021.

\bibitem{feng2023rethinking}
Qi Feng, Kun He, He Wen, Cem Keskin, and Yuting Ye.
\newblock Rethinking the data annotation process for multi-view 3d pose estimation with active learning and self-training.
\newblock In {\em WACV}, 2023.

\bibitem{AL4HPE_RL}
Erik G{\"{a}}rtner, Aleksis Pirinen, and Cristian Sminchisescu.
\newblock Deep reinforcement learning for active human pose estimation.
\newblock In {\em AAAI}, 2020.

\bibitem{pose_icsports}
Brennan Gebotys, Alexander Wong, and David~A. Clausi.
\newblock {POOF:} efficient goalie pose annotation using optical flow.
\newblock In {\em icSPORTS}, 2021.

\bibitem{DetTrack}
Rohit Girdhar, Georgia Gkioxari, Lorenzo Torresani, Manohar Paluri, and Du Tran.
\newblock Detect-and-track: Efficient pose estimation in videos.
\newblock In {\em CVPR}, 2018.

\bibitem{ICCV_AE_Anomaly}
Dong Gong, Lingqiao Liu, Vuong Le, Budhaditya Saha, Moussa~Reda Mansour, Svetha Venkatesh, and Anton van~den Hengel.
\newblock Memorizing normality to detect anomaly: Memory-augmented deep autoencoder for unsupervised anomaly detection.
\newblock In {\em ICCV}, 2019.

\bibitem{AL4HPE_MATAL}
Jia Gong, Zhipeng Fan, Qiuhong Ke, Hossein Rahmani, and Jun Liu.
\newblock Meta agent teaming active learning for pose estimation.
\newblock In {\em CVPR}, 2022.

\bibitem{ASD1}
Daniel Groos, Lars Adde, Ragnhild St{\o}en, Heri Ramampiaro, and Espen A.~F. Ihlen.
\newblock Towards human-level performance on automatic pose estimation of infant spontaneous movements.
\newblock {\em Comput. Med. Imaging Graphics}, 95:102012, 2022.

\bibitem{DeepAL}
Kuan{-}Hao Huang.
\newblock Deepal: Deep active learning in python.
\newblock {\em arXiv preprint arXiv:2111.15258}, 2021.

\bibitem{ActiveImageSegmentationPropagation}
Suyog~Dutt Jain and Kristen Grauman.
\newblock Active image segmentation propagation.
\newblock In {\em CVPR}, 2016.

\bibitem{OTPose}
Kyung{-}Min Jin, Gun{-}Hee Lee, and Seong{-}Whan Lee.
\newblock Otpose: Occlusion-aware transformer for pose estimation in sparsely-labeled videos.
\newblock In {\em SMC}, 2022.

\bibitem{DBLP:journals/cviu/KawanaUHY18}
Yuki Kawana, Norimichi Ukita, Jia{-}Bin Huang, and Ming{-}Hsuan Yang.
\newblock Ensemble convolutional neural networks for pose estimation.
\newblock {\em Comput. Vis. Image Underst.}, 169:62--74, 2018.

\bibitem{unc_aleatoric}
Alex Kendall and Yarin Gal.
\newblock What uncertainties do we need in bayesian deep learning for computer vision?
\newblock In {\em NIPS}, 2017.

\bibitem{Adam}
Diederik~P. Kingma and Jimmy Ba.
\newblock Adam: {A} method for stochastic optimization.
\newblock In {\em ICLR}, 2015.

\bibitem{Openpifpaf}
Sven Kreiss, Lorenzo Bertoni, and Alexandre Alahi.
\newblock Openpifpaf: Composite fields for semantic keypoint detection and spatio-temporal association.
\newblock {\em {IEEE} Trans. Intell. Transp. Syst.}, 23(8):13498--13511, 2022.

\bibitem{sc_namedentity}
Florian Laws and Hinrich Sch{\"{u}}tze.
\newblock Stopping criteria for active learning of named entity recognition.
\newblock In {\em COLING}, 2008.

\bibitem{HP}
David~D. Lewis and Jason Catlett.
\newblock Heterogeneous uncertainty sampling for supervised learning.
\newblock In {\em Mach. Learn.}, pages 148--156, 1994.

\bibitem{HPE_Testtime}
Yizhuo Li, Miao Hao, Zonglin Di, Nitesh~B. Gundavarapu, and Xiaolong Wang.
\newblock Test-time personalization with a transformer for human pose estimation.
\newblock In {\em NeurIPS}, 2021.

\bibitem{MSCOCO}
Tsung{-}Yi Lin, Michael Maire, Serge~J. Belongie, James Hays, Pietro Perona, Deva Ramanan, Piotr Doll{\'{a}}r, and C.~Lawrence Zitnick.
\newblock Microsoft {COCO:} common objects in context.
\newblock In {\em ECCV}, 2014.

\bibitem{AL_for_HPE}
Buyu Liu and Vittorio Ferrari.
\newblock Active learning for human pose estimation.
\newblock In {\em ICCV}, 2017.

\bibitem{TLforHPE_driver}
Yazhou Liu, Pongsak Lasang, Sugiri Pranata, Shengmei Shen, and Wenchao Zhang.
\newblock Driver pose estimation using recurrent lightweight network and virtual data augmented transfer learning.
\newblock {\em {IEEE} Trans. Intell. Transp. Syst.}, 20(10):3818--3831, 2019.

\bibitem{DCPose}
Zhenguang Liu, Haoming Chen, Runyang Feng, Shuang Wu, Shouling Ji, Bailin Yang, and Xun Wang.
\newblock Deep dual consecutive network for human pose estimation.
\newblock In {\em CVPR}, 2021.

\bibitem{AdamW}
Ilya Loshchilov and Frank Hutter.
\newblock Decoupled weight decay regularization.
\newblock In {\em ICLR}, 2019.

\bibitem{pose_anomaly}
Amir Markovitz, Gilad Sharir, Itamar Friedman, Lihi Zelnik{-}Manor, and Shai Avidan.
\newblock Graph embedded pose clustering for anomaly detection.
\newblock In {\em CVPR}, 2020.

\bibitem{DBLP:conf/mva/MatsumotoSMMMMU19}
Takuya Matsumoto, Kodai Shimosato, Takahiro Maeda, Tatsuya Murakami, Koji Murakoso, Kazuhiko Mino, and Norimichi Ukita.
\newblock Automatic human pose annotation for loose-fitting clothes.
\newblock In {\em MVA}, 2019.

\bibitem{ALforHPE_temporal_continuity}
Taro Mori, Daisuke Deguchi, Yasutomo Kawanishi, Ichiro Ide, Hiroshi Murase, and Tetsuo Inoshita.
\newblock {Active learning for human pose estimation based on temporal pose continuity}.
\newblock In {\em IWAIT}, 2022.

\bibitem{Hybrid_feature}
Bharath~Raj N., Anand Subramanian, Kashyap Ravichandran, and N. Venkateswaran.
\newblock Exploring techniques to improve activity recognition using human pose skeletons.
\newblock In {\em HADCV (WACVW)}, 2020.

\bibitem{densmap}
Ashwin Narayan, Bonnie Berger, and Hyunghoon Cho.
\newblock Assessing single-cell transcriptomic variability through density-preserving data visualization.
\newblock {\em Nat. Biotechnol.}, 39:765 -- 774, 2021.

\bibitem{HPE_in_bed}
Shunsuke Ochi and Jun Miura.
\newblock Depth-based in-bed human pose estimation with synthetic dataset generation and deep keypoint estimation.
\newblock In Leonid Karlinsky, Tomer Michaeli, and Ko Nishino, editors, {\em ACVR (ECCVW)}, 2022.

\bibitem{Seq-UPS}
Gaurav Patel, Jan~P. Allebach, and Qiang Qiu.
\newblock Seq-ups: Sequential uncertainty-aware pseudo-label selection for semi-supervised text recognition.
\newblock In {\em WACV}, 2023.

\bibitem{Hybrid_CenterOfGravity}
Marina Pismenskova, Oxana Balabaeva, Viacheslav Voronin, and Valentin Fedosov.
\newblock Classification of a two-dimensional pose using a human skeleton.
\newblock {\em MATEC Web Conf.}, 132:05016, 2017.

\bibitem{SemSeg_with_ActiveSemi-SupervisedLearning}
Aneesh Rangnekar, Christopher Kanan, and Matthew~J. Hoffman.
\newblock Semantic segmentation with active semi-supervised learning.
\newblock In {\em WACV}, 2023.

\bibitem{ALSurvey}
Pengzhen Ren, Yun Xiao, Xiaojun Chang, Po{-}Yao Huang, Zhihui Li, Brij~B. Gupta, Xiaojiang Chen, and Xin Wang.
\newblock A survey of deep active learning.
\newblock {\em {ACM} Comput. Surv.}, 54(9):1--40, 2022.

\bibitem{coreset}
Ozan Sener and Silvio Savarese.
\newblock Active learning for convolutional neural networks: {A} core-set approach.
\newblock In {\em ICLR}, 2018.

\bibitem{VL4Pose}
Megh Shukla, Roshan Roy, Pankaj Singh, Shuaib Ahmed, and Alexandre Alahi.
\newblock Vl4pose: Active learning through out-of-distribution detection for pose estimation.
\newblock In {\em BMVC}, 2022.

\bibitem{Hybrid_Angle}
Amarjot Singh, Devendra Patil, and S.~N. Omkar.
\newblock Eye in the sky: Real-time drone surveillance system {(DSS)} for violent individuals identification using scatternet hybrid deep learning network.
\newblock In {\em ECV (CVPRW)}, 2018.

\bibitem{HRNet_forHPE}
Ke Sun, Bin Xiao, Dong Liu, and Jingdong Wang.
\newblock Deep high-resolution representation learning for human pose estimation.
\newblock In {\em CVPR}, 2019.

\bibitem{DBLP:conf/iccv/UkitaHK09}
Norimichi Ukita, Michiro Hirai, and Masatsugu Kidode.
\newblock Complex volume and pose tracking with probabilistic dynamical models and visual hull constraints.
\newblock In {\em ICCV}, 2009.

\bibitem{DBLP:conf/eccv/UkitaTK08}
Norimichi Ukita, Ryosuke Tsuji, and Masatsugu Kidode.
\newblock Real-time shape analysis of a human body in clothing using time-series part-labeled volumes.
\newblock In {\em ECCV}, 2008.

\bibitem{DBLP:journals/cviu/UkitaU18}
Norimichi Ukita and Yusuke Uematsu.
\newblock Semi- and weakly-supervised human pose estimation.
\newblock {\em Comput. Vis. Image Underst.}, 170:67--78, 2018.

\bibitem{JRDB-Pose}
Edward Vendrow, Duy{-}Tho Le, Jianfei Cai, and Hamid Rezatofighi.
\newblock Jrdb-pose: {A} large-scale dataset for multi-person pose estimation and tracking.
\newblock In {\em CVPR}, 2023.

\bibitem{sc_foral}
Andreas Vlachos.
\newblock A stopping criterion for active learning.
\newblock {\em Comput. Speech Lang.}, 22(3):295--312, 2008.

\bibitem{SimpleBaseline}
Bin Xiao, Haiping Wu, and Yichen Wei.
\newblock Simple baselines for human pose estimation and tracking.
\newblock In {\em ECCV}, 2018.

\bibitem{DeciWatch}
Ailing Zeng, Xuan Ju, Lei Yang, Ruiyuan Gao, Xizhou Zhu, Bo Dai, and Qiang Xu.
\newblock Deciwatch: {A} simple baseline for 10{\(^\times\)} efficient 2d and 3d pose estimation.
\newblock In {\em ECCV}, 2022.

\bibitem{K-FPN}
Yuexi Zhang, Yin Wang, Octavia~I. Camps, and Mario Sznaier.
\newblock Key frame proposal network for efficient pose estimation in videos.
\newblock In {\em ECCV}, 2020.

\bibitem{kmeans}
Fedor Zhdanov.
\newblock Diverse mini-batch active learning.
\newblock {\em arXiv preprint arXiv:1901.05954}, 2019.

\bibitem{ATL_medicalimage}
Zongwei Zhou, Jae~Y. Shin, Suryakanth~R. Gurudu, Michael~B. Gotway, and Jianming Liang.
\newblock Active, continual fine tuning of convolutional neural networks for reducing annotation efforts.
\newblock {\em Medical Image Anal.}, 71:101997, 2021.

\bibitem{sc_zhu}
Jingbo Zhu, Huizhen Wang, and Eduard~H. Hovy.
\newblock Learning a stopping criterion for active learning for word sense disambiguation and text classification.
\newblock In {\em IJCNLP}, 2008.

\bibitem{sc_multi}
Jingbo Zhu, Huizhen Wang, and Eduard~H. Hovy.
\newblock Multi-criteria-based strategy to stop active learning for data annotation.
\newblock In {\em COLING}, 2008.

\end{thebibliography}
}

\clearpage


\appendix

\renewcommand{\thesection}{\Alph{section}}
\newcommand{\appendixhead}%
{\textbf{\huge Appendix}
\vspace{0.25in}}

\appendixhead

This appendix covers the additional results and visualizations that were excluded from the main paper due to a lack of space.
Specifically, the following contents are included: i) The complete table of our experiment results in the main paper, the impact of hyperparameter changes in our proposed method, and the validation of proposed uncertainty criteria (Sec.~\ref{sup:posetrack21}). ii) The experimental results of video-specific ATL on JRDB-Pose (Sec.~\ref{sup:jrdb-pose}), iii) A statement about the limitations of the proposed method (Sec.~\ref{sup:limit}), and iv) Additional qualitative examples of sample selection by proposed THC, WPU, and DUW (Sec.~\ref{sup:visualization}).

\begin{table*}[t]
  \caption{Quantitative results of our proposed video-specific ATL on PoseTrack21~\cite{PoseTrack21}. \red{Red} and \blue{blue} indicate the best and the second best, respectively. AP@0.6 is the average AP of 170 test videos with a 0.6 OKS threshold. ``5\%'' means the estimation result with 5\% labeled samples in the query video. ALC values are also calculated by an average of 170 test videos.}
    \centering
        \begin{tabular}{l| c c c c c c c c c c | c}
        \hline
        \multicolumn{1}{l|}{\multirow{2}{*}{Criterion}} & \multicolumn{10}{c}{\makebox[12mm]{AP@0.6 (\%)}} & \multicolumn{1}{|c}{ALC} \\
        \multicolumn{1}{l|}{} & 0\% & 5\% & 10\% & 15\% & 20\% & 30\% & 40\% & 60\% & 80\% & 100\% & \multicolumn{1}{|c}{(\%)} \\
        \hline
        Random & \textbf{\red{81.82}} & 87.76 & 93.60 & 95.10 & 96.09 & 96.92 & 97.39 & 98.50 & 99.21 & \textbf{\red{100.00}} & 96.91\\
        LC~\cite{HP} & \textbf{\red{81.82}} & 77.49 & 89.55 & 93.03 & 94.60 & 95.69 & 96.77 & 98.29 & 99.37 & \textbf{\red{100.00}} & 95.74 \\
        MPE~\cite{AL_for_HPE} & \textbf{\red{81.82}} & 78.96 & 90.96 & 93.98 & 95.09 & 96.44 & 97.23 & 98.38 & 99.28 & \textbf{\red{100.00}} & 96.11 \\
        TPC~\cite{ALforHPE_temporal_continuity} & \textbf{\red{81.82}} & 83.38 & 90.97 & 93.63 & 95.32 & 96.34 & 97.31 & 98.54 & 99.43 & \textbf{\red{100.00}} & 96.40 \\
        k-means~\cite{kmeans} & \textbf{\red{81.82}} & \textbf{\red{93.97}} & 95.19 & 95.82 & 96.37 & 97.55 & 98.11 & 98.82 & \textbf{\blue{99.45}} & \textbf{\red{100.00}} & 97.65 \\
        Core-Set~\cite{coreset} & \textbf{\red{81.82}} & 93.18 & \textbf{\red{96.35}} & \textbf{\blue{97.26}} & \textbf{\blue{97.62}} & \textbf{\blue{98.18}} & \textbf{\blue{98.60}} & \textbf{\blue{99.27}} & \textbf{\red{99.67}} & \textbf{\red{100.00}} & \textbf{\blue{98.12}} \\
        \begin{tabular}[c]{@{}l@{}}\textbf{Ours}\\ \textbf{(THC+WPU+DUW)}\end{tabular} & \textbf{\red{81.82}} & \textbf{\blue{93.35}} & \textbf{\blue{96.14}} & \textbf{\red{97.37}} & \textbf{\red{97.90}} & \textbf{\red{98.44}} & \textbf{\red{98.77}} & \textbf{\red{99.33}} & \textbf{\red{99.67}} & \textbf{\red{100.00}} & \textbf{\red{98.21}} \\
        \hline
        \end{tabular}
    \centering
    \label{table:PoseTrack21_full}
\end{table*}

\section{Additional Results on PoseTrack21}
\label{sup:posetrack21}
In this section, we provide a complete table showing the experimental results for the video-specific Active Transfer Learning (ATL) on the PoseTrack21 dataset~\cite{PoseTrack21}, which could not be included in the main paper. All the results are obtained with the same experimental settings as described in the main manuscript. 
Once again, the selection criteria used in our experiment are as follows:

\begin{itemize}[noitemsep, topsep=1pt, parsep=1pt, partopsep=1pt]
    \item \textbf{Random:} Random sampling from a uniform distribution.
    \item \textbf{Least Confidence (LC):} A traditional uncertainty measurement described in~\cite{HP}.
    \item \textbf{Multiple Peak Entropy (MPE):} An uncertainty criterion in~\cite{AL_for_HPE}.
    \item \textbf{Temporal Pose Continuity (TPC):} An uncertainty criterion in~\cite{ALforHPE_temporal_continuity}.
    \item \textbf{k-means:} A representativeness criterion used in~\cite{kmeans}.
    \item \textbf{Core-Set:} An original Core-Set sampling from~\cite{coreset}.
    \item \textbf{Temporal Heatmap Continuity (THC):} Our uncertainty criterion based on the temporal change of estimated heatmaps.
    \item \textbf{Whole-body Pose Unnaturalness (WPU):} Our uncertainty criterion based on the unnaturalness of estimated poses.
    \item \textbf{Dynamic Uncertainty Weighting (DUW):} Our criterion combines uncertainty and Core-Set sampling~\cite{coreset}.
\end{itemize}

\subsection{Baseline and State-of-the-art Comparisons}
\label{sub:sota}
Table~\ref{table:PoseTrack21_full} shows full results of the proposed video-specific ATL on PoseTrack21~\cite{PoseTrack21}. While the performances of all uncertainty-based methods~\cite{HP, AL_for_HPE, ALforHPE_temporal_continuity} are even less than the random selection, our proposed method (THC+WPU+DUW) stably outperforms other methods including k-means~\cite{kmeans} and Core-Set~\cite{coreset}. Our proposed method demonstrates high performance consistently. However, during the initial cycles of ATL, representativeness criteria such as k-means clustering~\cite{kmeans} and Core-Set sampling~\cite{coreset} show superior performance. This observation aligns with our hypothesis stated in the main paper: during the early stages of ATL, it is important to cover the data distribution of the target domain. In contrast, as ATL progresses, identifying samples with high uncertainty becomes increasingly important. In this context, our DUW effectively enhances performance by identifying these challenging samples. This shows the importance of uncertainty for performance improvement during the later stages of ATL.

\begin{table*}[t]
  \caption{Ablation study results of video-specific ATL on PoseTrack21~\cite{PoseTrack21}. \red{Red} and \blue{blue} indicate the best and the second best, respectively. AP@0.6 is the average AP of 170 test videos with a 0.6 OKS threshold. ``5\%'' means the estimation result with 5\% labeled samples. ALC values are also calculated by an average of 170 test videos. (fixed), (increase), (const), and (decrease) denote the video-specific ATL with a fixed balance of uncertainty and representativeness, linearly increasing/decreasing the weight of THC toward WPU, using the same weight for THC and WPU, respectively. }
  \centering
    \begin{tabular}{l| c c c c c c c c c c | c}
    \hline
    \multicolumn{1}{l|}{\multirow{2}{*}{Criterion}} & \multicolumn{10}{c}{\makebox[12mm]{AP@0.6 (\%)}} & \multicolumn{1}{|c}{\makebox[8mm]{ALC}} \\
    \multicolumn{1}{l|}{} & 0\% & 5\% & 10\% & 15\% & 20\% & 30\% & 40\% & 60\% & 80\% & 100\% & \multicolumn{1}{|c}{(\%)} \\
    \hline
    Core-Set~\cite{coreset} & \textbf{\red{81.82}} & 93.18 & \textbf{\red{96.35}} & 97.26 & 97.62 & 98.18 & 98.60 & 99.27 & 99.67 & \textbf{\red{100.00}} & 98.12 \\
    \hline
    THC & \textbf{\red{81.82}} & 82.59 & 89.10 & 91.85 & 92.86 & 94.70 & 96.43 & 97.74 & 98.97 & \textbf{\red{100.00}} & 95.45 \\
    WPU & \textbf{\red{81.82}} & 85.56 & 91.11 & 93.39 & 94.74 & 96.36 & 97.31 & 98.48 & 99.28 & \textbf{\red{100.00}} & 96.45 \\
    THC+WPU & \textbf{\red{81.82}} & 84.82 & 91.72 & 93.83 & 95.17 & 96.38 & 97.25 & 98.54 & 99.35 & \textbf{\red{100.00}} & 96.51 \\
    THC+DUW & \textbf{\red{81.82}} & 93.12 & 95.88 & 97.11 & 97.70 & 98.42 & \textbf{\blue{98.91}} & \textbf{\blue{99.35}} & \textbf{\red{99.74}} & \textbf{\red{100.00}} & 98.19 \\
    WPU+DUW & \textbf{\red{81.82}} & \textbf{\blue{93.19}} & 95.96 & 97.26 & \textbf{\blue{97.87}} & \textbf{\red{98.51}} & 98.76 & 99.27 & 99.65 & \textbf{\red{100.00}} & 98.17 \\
    \hline
    THC+WPU+DUW & \multirow{2}{*}{\textbf{\red{81.82}}} & \multirow{2}{*}{93.02} & \multirow{2}{*}{95.71} & \multirow{2}{*}{97.15} & \multirow{2}{*}{97.68} & \multirow{2}{*}{98.41} & \multirow{2}{*}{98.81} & \multirow{2}{*}{99.28} & \multirow{2}{*}{99.66} & \multirow{2}{*}{\textbf{\red{100.00}}} & \multirow{2}{*}{98.14} \\
    (fixed) & & & & & & & & & & & \\
    \hline
    THC+WPU+DUW & \multirow{2}{*}{\textbf{\red{81.82}}} & \multirow{2}{*}{93.18} & \multirow{2}{*}{95.85} & \multirow{2}{*}{97.13} & \multirow{2}{*}{97.86} & \multirow{2}{*}{98.47} & \multirow{2}{*}{98.80} & \multirow{2}{*}{99.27} & \multirow{2}{*}{99.64} & \multirow{2}{*}{\textbf{\red{100.00}}} & \multirow{2}{*}{98.16} \\
    (increase) & & & & & & & & & & & \\
    THC+WPU+DUW & \multirow{2}{*}{\textbf{\red{81.82}}} & \multirow{2}{*}{93.08} & \multirow{2}{*}{96.11} & \multirow{2}{*}{\textbf{\blue{97.34}}} & \multirow{2}{*}{97.72} & \multirow{2}{*}{\textbf{\blue{98.48}}} & \multirow{2}{*}{\textbf{\red{98.94}}} & \multirow{2}{*}{\textbf{\red{99.43}}} & \multirow{2}{*}{\textbf{\blue{99.73}}} & \multirow{2}{*}{\textbf{\red{100.00}}} & \multirow{2}{*}{\textbf{\red{98.24}}} \\
    (decrease) & & & & & & & & & & &\\
    THC+WPU+DUW & \multirow{2}{*}{\textbf{\red{81.82}}} & \multirow{2}{*}{\textbf{\red{93.35}}} & \multirow{2}{*}{\textbf{\blue{96.14}}} & \multirow{2}{*}{\textbf{\red{97.37}}} & \multirow{2}{*}{\textbf{\red{97.90}}} & \multirow{2}{*}{98.44} & \multirow{2}{*}{98.77} & \multirow{2}{*}{99.33} & \multirow{2}{*}{99.67} & \multirow{2}{*}{\textbf{\red{100.00}}} & \multirow{2}{*}{\textbf{\blue{98.21}}} \\
    (const) & & & & & & & & & & & \\
    \hline
    \end{tabular}
    \label{table:ablation_full}
\end{table*}

\subsection{Ablation Studies}
\label{sub:ablation}
Table~\ref{table:ablation_full} shows the full results of ablation studies. Our proposed methods, THC, WPU, and DUW, all used together, achieved the highest ALC. THC+DUW and WPU+DUW surpassed the performance of the original Core-Set~\cite{coreset} due to the incorporation of uncertainty in sample selection.
In cases where only THC, only WPU, or THC+WPU, the performance is found to be inferior to that of the Core-Set~\cite{coreset}. These lower performances are attributed to a selection bias~\cite{ALSurvey, al_bias} common in sample selection by uncertainty.

In addition, as mentioned in Sec.~\ref{exp:ablation} of the main paper, we investigated the performance of ATL under various configurations conceivable for the proposed method's components. As shown in Table~\ref{table:ablation_full}, without dynamically combining uncertainty and representativeness at each stage of active learning (``fixed''), the performance was not as promising compared to other configurations of our proposed method.

Furthermore, regarding the combination of THC and WPU, generally, the use of THC and WPU with the same weight (``const'') led to significant performance improvements in the initial phase of active learning. On the other hand, by gradually decreasing the weight of THC and increasing the weight of WPU (``decrease''), performance efficiently improved from the mid-phase of ATL, resulting in the highest value for ALC. This suggests that not only high performance can be achieved by simply using THC and WPU with the same weight (``const''), but also designing an appropriate method for combining THC and WPU can lead to a more effective video-specific ATL.

\subsection{Impact of Hyperparameter Changes in DUW}
\label{sup:hyper}
In this section, we investigate the influence of the hyperparameter $\lambda$ in the following objective function used in DUW:

\begin{equation}
     u = \argmax_{i \in U} \{\min_{j \in L} \{(1-G_c) \times \Delta(x_i, x_j)\} + G_c \times \lambda C(x_i)\}
\end{equation}

where $\Delta(x_i, x_j)$ is the Euclidean distance between the sample's feature vectors $x_i$ and $x_j$, $G_c$ is the approximated generalization performance of Human Pose (HP) estimator, $C(x_i)$ is each sample's uncertainty score, $L$ represents labeled samples set, and $U$ represents an unlabeled set, respectively.
As explained in the main manuscript, a larger value of $\lambda$ leads to a sample selection that emphasizes uncertainty more. Conversely, a smaller value of $\lambda$ results in a sample selection more similar to the original Core-Set sampling~\cite{coreset}. When $\lambda$ equals zero, the sample selection is the same as the original Core-Set sampling.

\begin{table*}[t]
  \caption{Impact of variation in hyperparameter $\lambda$ of DUW. \red{Red} and \blue{blue} indicate the best and the second best, respectively. AP@0.6 is the average AP of 170 test videos with a 0.6 OKS threshold. ``5\%'' means the estimation result with 5\% labeled samples. ALC values are also calculated by an average of 170 test videos. When $\lambda = 0$, the sample selection is equivalent to the original Core-Set sampling~\cite{coreset}.}
  \centering
    \begin{tabular}{l| c c c c c c c c c c | c}
    \hline
    \multicolumn{1}{l|}{\multirow{2}{*}{Criterion}} & \multicolumn{10}{c}{\makebox[12mm]{AP@0.6 (\%)}} & \multicolumn{1}{|c}{\makebox[8mm]{ALC}} \\
    \multicolumn{1}{l|}{} & 0\% & 5\% & 10\% & 15\% & 20\% & 30\% & 40\% & 60\% & 80\% & 100\% & \multicolumn{1}{|c}{(\%)} \\
    \hline
    \begin{tabular}[c]{@{}l@{}}$\lambda=0$\\ (Core-Set~\cite{coreset})\end{tabular} & \textbf{\red{81.82}} & 93.18 & \textbf{\red{96.35}} & 97.26 & 97.62 & 98.18 & 98.60 & 99.27 & \textbf{\blue{99.67}} & \textbf{\red{100.00}} & 98.12 \\
    $\lambda=0.001$ & \textbf{\red{81.82}} & 93.30 & \textbf{\blue{96.33}} & \textbf{\blue{97.34}} & 97.78 & \textbf{\red{98.50}} & \textbf{\red{98.82}} & \textbf{\blue{99.28}} & 99.62 & \textbf{\red{100.00}} & \textbf{\blue{98.20}} \\
    $\lambda=0.01$ & \textbf{\red{81.82}} & \textbf{\blue{93.35}} & 96.14 & \textbf{\red{97.37}} & \textbf{\red{97.90}} & \textbf{\blue{98.44}} & 98.77 & \textbf{\red{99.33}} & \textbf{\blue{99.67}} & \textbf{\red{100.00}} & \textbf{\red{98.21}} \\
    $\lambda=0.1$ & \textbf{\red{81.82}} & \textbf{\red{93.37}} & \textbf{\red{96.35}} & 97.29 & \textbf{\blue{97.82}} & 98.36 & \textbf{\blue{98.80}} & 99.26 & \textbf{\red{99.70}} & \textbf{\red{100.00}} & \textbf{\blue{98.20}} \\
    $\lambda=1$ & \textbf{\red{81.82}} & 93.31 & 95.77 & 97.05 & 97.55 & 98.10 & 98.43 & 99.12 & 99.60 & \textbf{\red{100.00}} & 98.01 \\
    $\lambda=10$ & \textbf{\red{81.82}} & 93.34 & 95.64 & 96.64 & 97.12 & 97.99 & 98.36 & 99.07 & 99.58 & \textbf{\red{100.00}} & 97.91 \\
    \hline
    \end{tabular}
    \label{table:hyperparameter}
\end{table*}

Table~\ref{table:hyperparameter} shows the results of video-specific ATL when the order of $\lambda$ in Eq.(\ref{eq:duw}) is changed. The active selection criterion used for this experiment was THC+WPU+DUW, and the detailed experimental setup remains the same as Sec.~\ref{sub:sota} and~\ref{sub:ablation}. According to the results presented in Table~\ref{table:hyperparameter}, the highest ALC is achieved when $\lambda$ equals 0.01, followed closely by $\lambda$ values of 0.1 and 0.001. On the other hand, performance is degraded when $\lambda$ equals 0, corresponding to the original Core-Set~\cite{coreset} that tends to select samples with low informativeness due to not considering uncertainty. Similarly, performance decreases with $\lambda$ values of 1 and 10, which overly prioritize uncertainty and consequently lose diversity in sample selection. 

\begin{figure}[t]
  \centering
  \begin{center}
    \includegraphics[clip, width=\columnwidth]{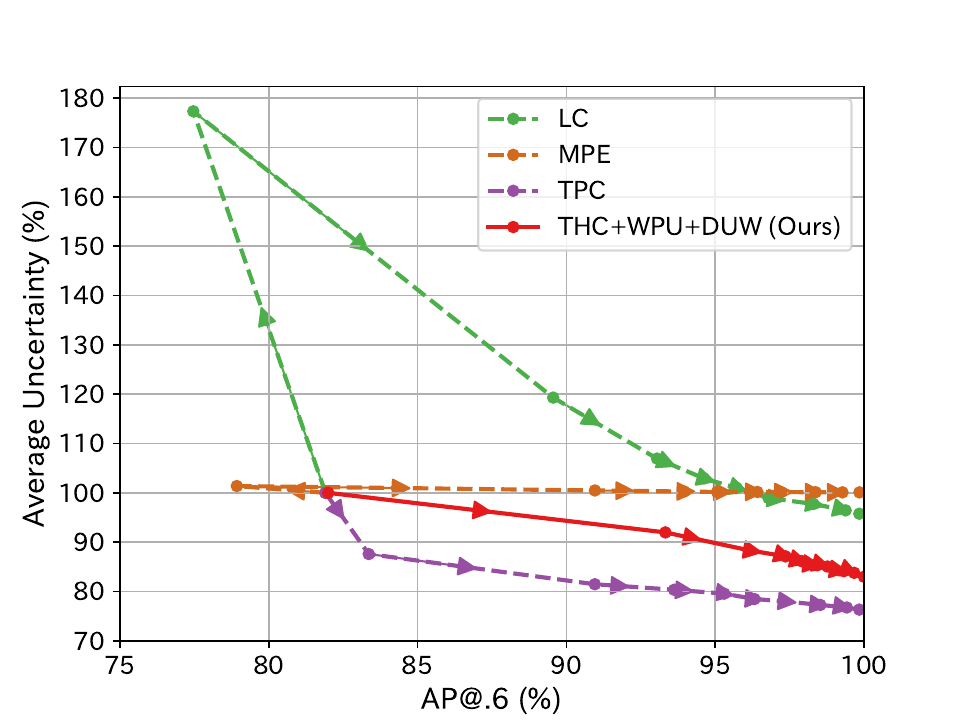}
  \end{center}
  \caption{The change in the average uncertainty accompanying the AP@0.6 transition in video-specific ATL on PoseTrack21~\cite{PoseTrack21}, which are shown in Tables~\ref{table:PoseTrack21_full} and~\ref{table:ablation_full}. The uncertainty at the beginning of ATL is used as a baseline (100\%).}
  \label{fig:uncertainty}
  \vspace{-0.2cm}
\end{figure}

\subsection{Validation of Uncertainty Criteria}
\label{exp:uncval}
We validate our proposed uncertainty criteria, THC and WPU. Uncertainty is expected to be low when HP estimation is correct and high when the result is incorrect. Hence, as the accuracy of estimation improves, the uncertainty should decrease. Fig.~\ref{fig:uncertainty} shows the uncertainty change accompanying the transition of HP estimation performance in video-specific ATL. In MPE~\cite{AL_for_HPE}, the uncertainty remains almost constant regardless of changes in performance. On the other hand, THC and WPU show a desirable change in uncertainty that decreases with the improvement of AP.


\begin{figure}[t]
  \centering
  \begin{center}
    \includegraphics[clip, width=\columnwidth]{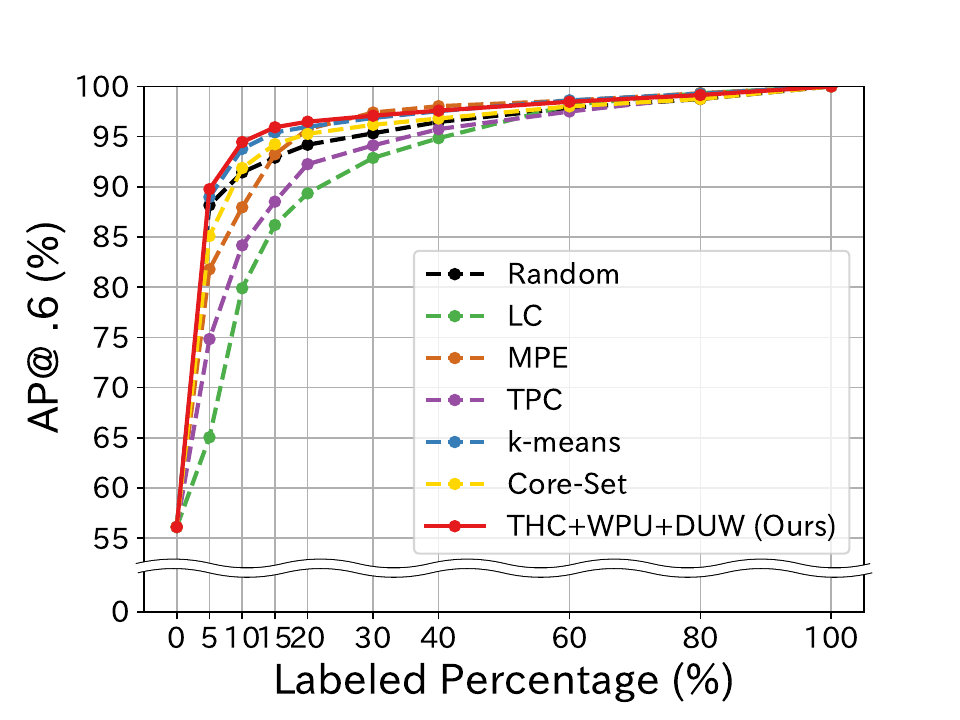}
  \end{center}
  \caption{Learning Curve of video-specific ATL on JRDB-Pose~\cite{JRDB-Pose}. }
  \label{figJRDBresult}
  \vspace{-0.2cm}
\end{figure}

\section{Additional Results on JRDB-Pose}
\label{sup:jrdb-pose}
In this section, we report a complete table of experimental results on the JRDB-Pose dataset~\cite{JRDB-Pose} in the main paper, additional experimental results with another metric, and the validation of uncertainty criteria on JRDB-Pose. The basic experimental conditions are the same as those described in Sec.~\ref{exp:evalandimpl} of the main paper.

\subsection{Baseline and State-of-the-art Comparison}
The results of the video-specific ATL for the 15 test videos from JRDB-Pose are presented in Tables~\ref{supptable:JRDBPose_OSPA} and~\ref{supptable:JRDB-Pose_full}. Fig.~\ref{figJRDBresult} is the plotted learning curve. Here too, while the initial AP@0.6 is 56.11\%, our proposed method (``Ours'') achieved performance close to 90\% with only 5\% of the labeling. Furthermore, our ALC performance across the entire ATL outperformed all comparison methods. 

Moreover, when evaluated using OSPA, our method displayed performance comparable to the best-performing method (``k-means~\cite{kmeans}''). This suggests that our method can efficiently adapt the HP estimator even for challenging datasets like JRDB-Pose~\cite{JRDB-Pose}.

\begin{table}[t]
  \caption{Quantitative results of our proposed video-specific ATL on JRDB-Pose~\cite{JRDB-Pose}. \red{Red} and \blue{blue} indicate the best and the second best, respectively. OSPA is an average of 15 test videos. ``5\%'' means the estimation result with 5\% labeled samples in the query video. ALC values are also an average of 15 test videos.}
  \centering
    \begin{center}
        \begin{tabular}{l| c c c | c}
        \hline
        \multicolumn{1}{l|}{\multirow{2}{*}{Criterion}} & \multicolumn{3}{c}{\makebox[12mm]{OSPA ↓}} & \multicolumn{1}{|c}{ALC ↓} \\
        \multicolumn{1}{l|}{} & 5\% & 20\% & 40\% & \multicolumn{1}{|c}{$\times10^-4$} \\
        \hline
        Random & \textbf{\red{0.129}} & 0.074 & \textbf{\blue{0.047}} & 5.22 \\
        LC~\cite{HP} & 0.287 & 0.122 & 0.068 & 7.80 \\
        MPE~\cite{AL_for_HPE} & 0.196 & 0.071 & \textbf{\red{0.040}} & 5.43 \\
        TPC~\cite{ALforHPE_temporal_continuity} & 0.239 & 0.110 & 0.066 & 7.59 \\
        k-means~\cite{kmeans} & \textbf{\blue{0.132}} & \textbf{\red{0.067}} & \textbf{\red{0.040}} & \textbf{\red{4.72}} \\
        Core-Set~\cite{coreset} & 0.157 & 0.077 & 0.052 & 5.70 \\
        \begin{tabular}[c]{@{}l@{}}\textbf{Ours}\\ \textbf{(THC+WPU+DUW)}\end{tabular} & 0.134 & \textbf{\blue{0.068}} & \textbf{\blue{0.047}} & \textbf{\blue{5.05}} \\
        \hline
        \end{tabular}
    \end{center}
    \label{supptable:JRDBPose_OSPA}
    \vspace{-0.2cm}
\end{table}

\begin{table*}[t]
  \caption{Quantitative results of our proposed video-specific ATL on JRDB-Pose~\cite{JRDB-Pose}. \red{Red} and \blue{blue} indicate the best and the second best, respectively. AP@0.6 is the average AP of 15 test videos with a 0.6 OKS threshold. ``5\%'' means the estimation result with 5\% labeled samples in the query video. ALC values are also calculated by an average of 15 test videos.}
    \centering
        \begin{tabular}{l| c c c c c c c c c c | c}
        \hline
        \multicolumn{1}{l|}{\multirow{2}{*}{Criterion}} & \multicolumn{10}{c}{\makebox[12mm]{AP@0.6 (\%)}} & \multicolumn{1}{|c}{ALC} \\
        \multicolumn{1}{l|}{} & 0\% & 5\% & 10\% & 15\% & 20\% & 30\% & 40\% & 60\% & 80\% & 100\% & \multicolumn{1}{|c}{(\%)} \\
        \hline
        Random & \textbf{\red{56.11}} & 88.16 & 91.44 & 92.91 & 94.19 & 95.33 & 96.46 & 97.86 & 98.74 & \textbf{\red{100.00}} & 95.42\\
        LC~\cite{HP} & \textbf{\red{56.11}} & 65.04 & 79.89 & 86.20 & 89.34 & 92.88 & 94.84 & 98.14 & \textbf{\red{99.32}} & \textbf{\red{100.00}} & 92.67 \\
        MPE~\cite{AL_for_HPE} & \textbf{\red{56.11}} & 81.78 & 87.95 & 93.24 & 95.74 & \textbf{\red{97.39}} & \textbf{\red{98.03}} & \textbf{\blue{98.59}} & \textbf{\red{99.32}} & \textbf{\red{100.00}} & 95.76 \\
        TPC~\cite{ALforHPE_temporal_continuity} & \textbf{\red{56.11}} & 74.83 & 84.16 & 88.52 & 92.25 & 94.12 & 95.74 & 97.50 & 98.94 & \textbf{\red{100.00}} & 93.76 \\
        k-means~\cite{kmeans} & \textbf{\red{56.11}} & \textbf{\blue{88.97}} & \textbf{\blue{93.78}} & \textbf{\blue{95.41}} & \textbf{\blue{95.98}} & 96.86 & 97.53 & \textbf{\red{98.61}} & \textbf{\blue{99.28}} & \textbf{\red{100.00}} & \textbf{\blue{96.41}} \\
        Core-Set~\cite{coreset} & \textbf{\red{56.11}} & 85.09 & 91.87 & 94.24 & 95.27 & 96.18 & 96.80 & 98.01 & 98.75 & \textbf{\red{100.00}} & 95.60 \\
        \begin{tabular}[c]{@{}l@{}}\textbf{Ours}\\ \textbf{(THC+WPU+DUW)}\end{tabular} & \textbf{\red{56.11}} & \textbf{\red{89.76}} & \textbf{\red{94.45}} & \textbf{\red{95.93}} & \textbf{\red{96.48}} & \textbf{\blue{97.08}} & \textbf{\blue{97.59}} & 98.47 & 99.14 & \textbf{\red{100.00}} & \textbf{\red{96.52}} \\
        \hline
        \end{tabular}
    \centering
    \label{supptable:JRDB-Pose_full}
\end{table*}

\subsection{Validation of Uncertainty Criteria}
\begin{figure}[t]
  \centering
  \begin{center}
    \includegraphics[clip, width=\columnwidth]{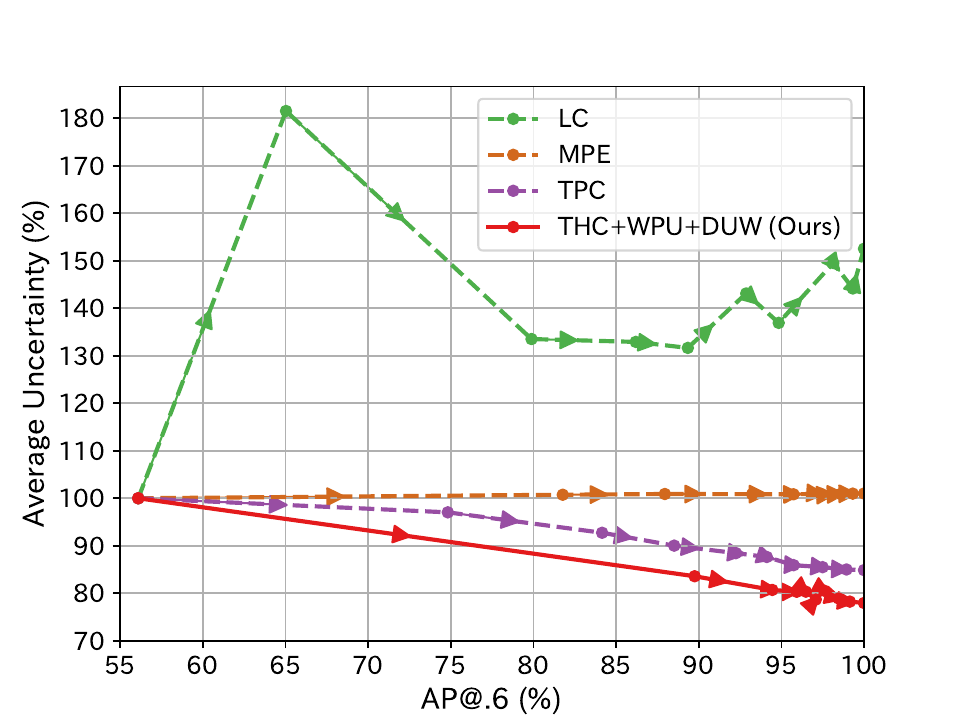}
  \end{center}
  \caption{The change in the average uncertainty accompanying the AP@0.6 transition in video-specific ATL on JRDB-Pose~\cite{JRDB-Pose}, which are shown in Table~\ref{supptable:JRDB-Pose_full}. The uncertainty at the beginning of ATL is used as a baseline (100\%).}
  \label{fig:JRDB_uncertainty}
  \vspace{-0.2cm}
\end{figure}

Figure~\ref{fig:JRDB_uncertainty} illustrates the change in uncertainty with the improvement of the FastPose's~\cite{AlphaPose} performance in the video-specific ATL using the JRDB-Pose dataset~\cite{JRDB-Pose}. As with the results in PoseTrack21~\cite{PoseTrack21} (Sec.~\ref{exp:uncval}), as the performance of FastPose improves, the value of uncertainty decreases. This suggests that our proposed uncertainty criteria (THC+WPU) can reflect the error of the prediction results.

\section{Limitations}
\label{sup:limit}
Despite the positive results observed in our study, it is important to acknowledge some limitations.

First, the tuning of learning conditions can be challenging. In particular, the optimal hyperparameters of the proposed methods could be dataset-specific and still require some tuning to achieve acceptable results.

Second, our Temporal Heatmap Continuity (THC) method has an inherent limitation. The THC might be high wrongly, especially when the object's movements are drastic. This could lead to a skewed selection towards instances with more intense movements.

Lastly, determining an upper bound for learning efficiency in ATL is still challenging. Mainly due to the too many cases of possible combinations of sample selections~\cite{K-FPN}, it is difficult to define an optimal sampling strategy during the ATL process even if ground truth labels are available.

These limitations offer potential areas to further improve the proposed method.

\section{Visualization of Proposed Active Selection Criteria}
\label{sup:visualization}
\begin{figure*}[t]
  \centering
  \begin{subfigure}{\linewidth}
    \centering
    \includegraphics[width=0.85\linewidth]{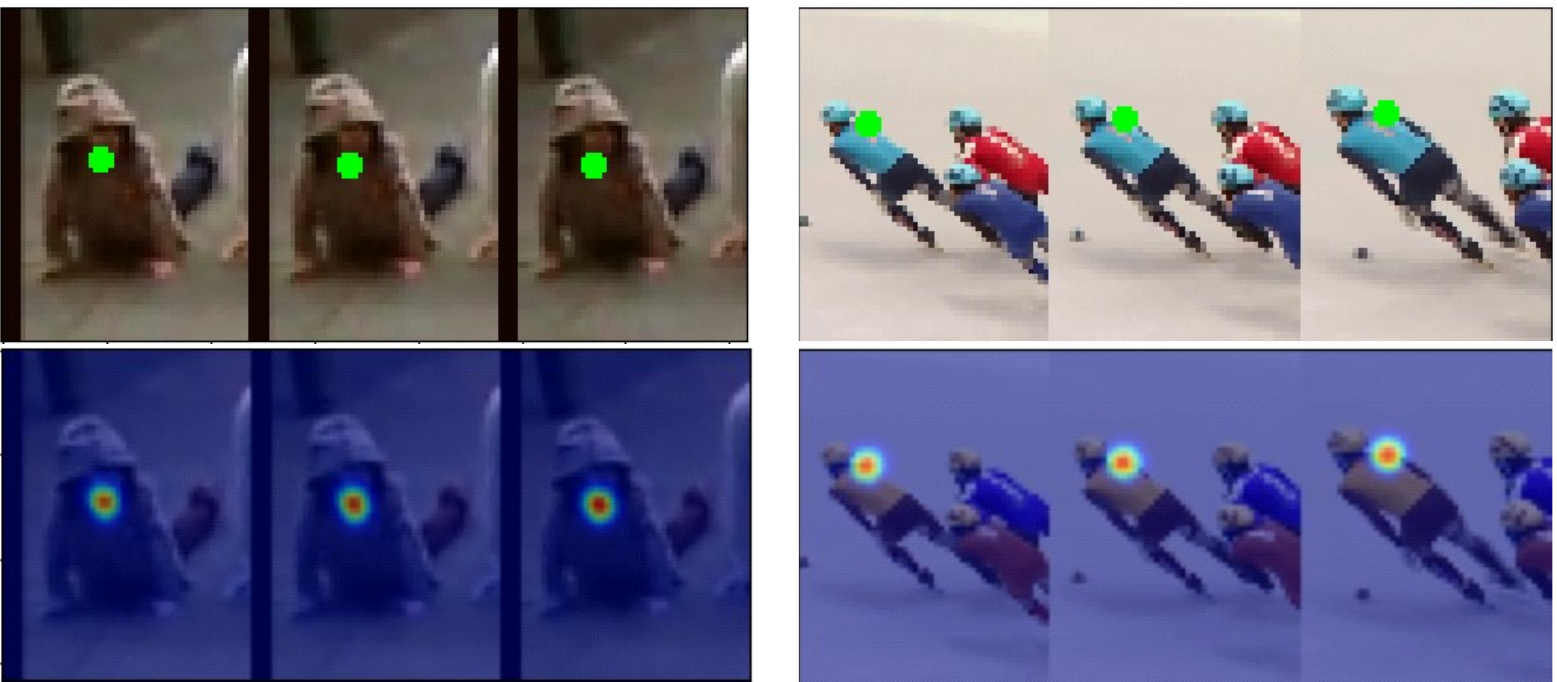}
    \caption{Heatmaps with low THC. Times are at t-1, t and t+1 from left to right in each scene.}
    \label{fig:low_thc_expand}
  \end{subfigure}
  \begin{subfigure}{\linewidth}
    \centering
    \includegraphics[width=0.85\linewidth]{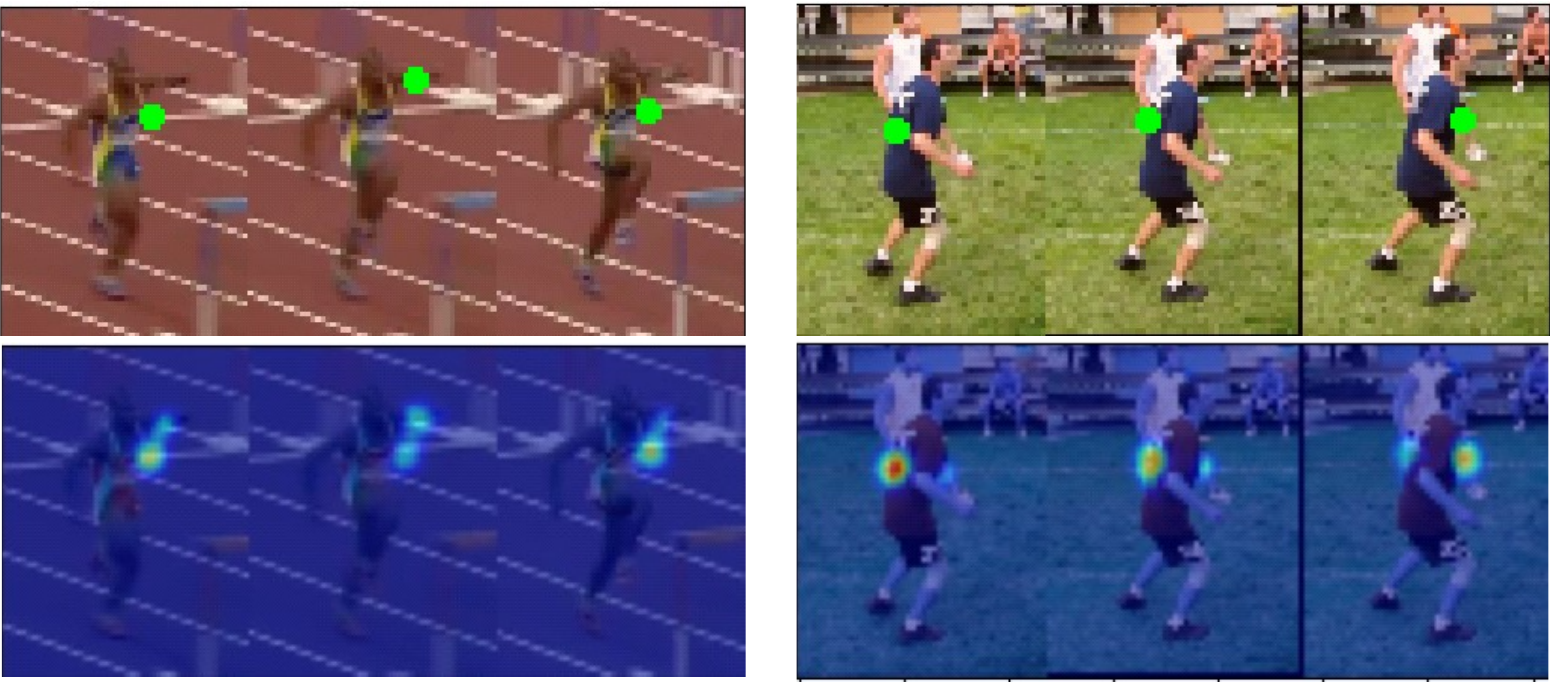}
    \caption{Heatmaps with high THC. Times are at t-1, t and t+1 from left to right in each scene.}
    \label{fig:high_thc_expand}
  \end{subfigure}
  \caption{Additional qualitative examples of our THC. The top row of the figure shows the original images, with the estimated keypoint positions marked by green circles. The bottom row presents the heatmaps estimated for each of the three adjacent frames, where a color closer to \blue{blue} indicates a lower probability of keypoint presence, while a color closer to \red{red} suggests a higher probability. (a) There is a strong peak at a single point in the heatmap between adjacent frames consistently. As a result, estimated keypoint positions are accurate. (b) In contrast, the estimations are inconsistent and the peaks in the heatmap are dispersed. It results in an erroneous estimation.}
  \label{fig:example_thc_expand}
\end{figure*}

\begin{figure*}[t]
  \centering
  \begin{subfigure}{\linewidth}
    \centering
    \includegraphics[width=\linewidth]{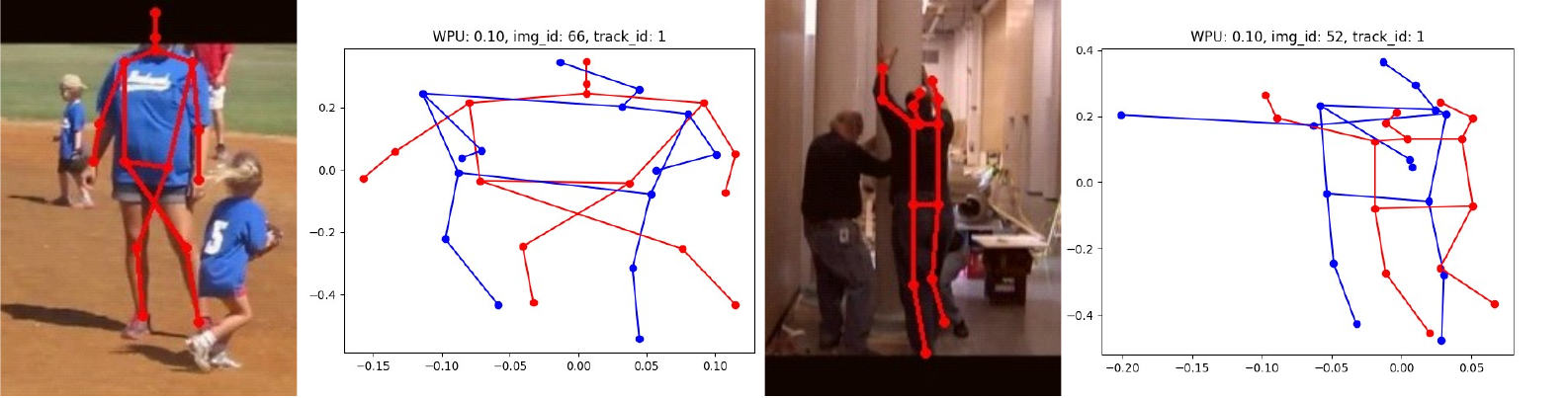}
    \caption{Estimated pose with low WPU.}
    \label{fig:low_wpu}
  \end{subfigure}
  \begin{subfigure}{\linewidth}
    \centering
    \includegraphics[width=\linewidth]{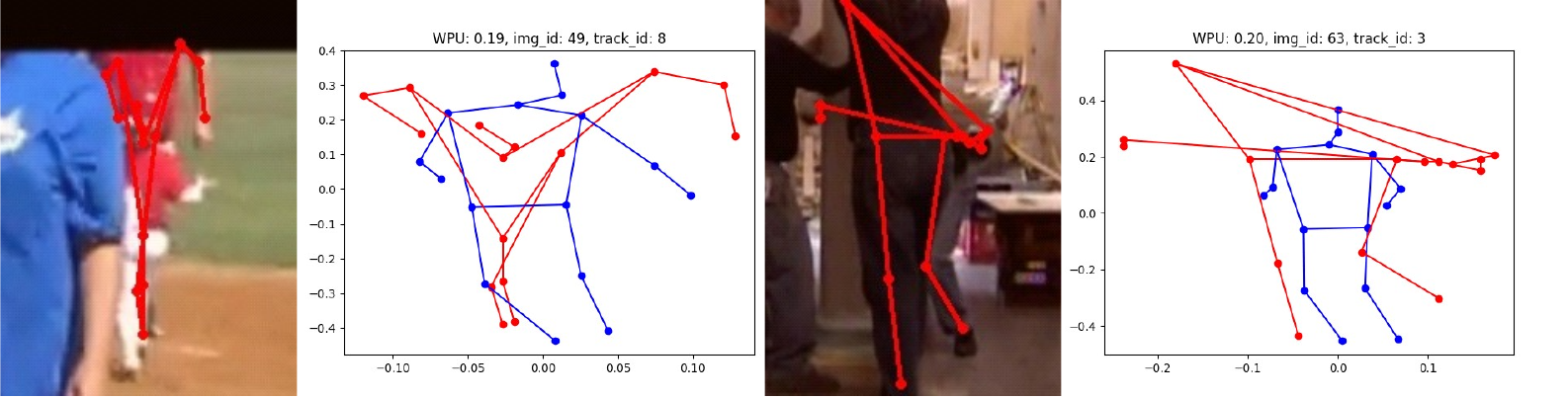}
    \caption{Estimated pose with high WPU.}
    \label{fig:high_wpu}
  \end{subfigure}
  \caption{Examples of samples selected by WPU. In the figure, the \red{red lines} represent the estimated pose and its Hybrid feature~\cite{Hybrid_feature}, and the \blue{blue lines} represent the Hybrid feature output by the AE trained on natural poses. In the case of (a), where the WPU is low, the \red{red} Hybrid feature, which is the input to the AE, and the \blue{blue} Hybrid feature, which is the output, are close to each other, and the estimated pose is also close to the correct one. On the other hand, in (b), due to an incorrect pose estimation input, the Hybrid features are far apart from each other, resulting in a high WPU value.}
  \label{fig:example_wpu_expand}
\end{figure*}

\begin{figure*}[t]
    \centering
    \includegraphics[width=\linewidth]{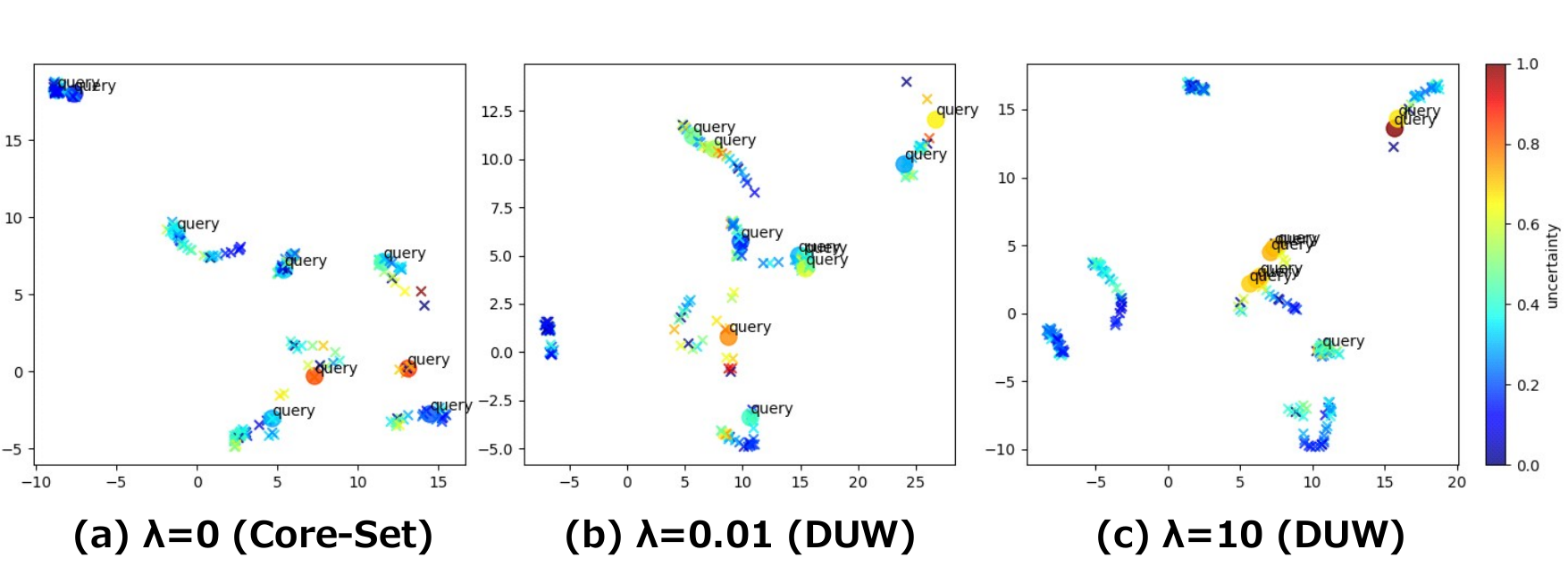}
    \caption{A visualization result of the sample selection of our DUW criterion. We have utilized DensMAP~\cite{densmap} to plot feature vectors extracted by the HP estimator. In this plot, circles represent newly selected samples, while cross marks denote unlabeled samples that were not selected for the current ATL cycle. The color of the plot corresponds to the normalized uncertainty. (a) represents the results when $\lambda=0$ (i.e., equivalent to the original Core-Set~\cite{coreset}), which tends to select diverse but uninformative samples with low uncertainty. (b) represents the selection for $\lambda=0.01$, which yielded the best results in Sec.~\ref{sup:hyper}. It can be seen that uncertain and diverse samples are selected. (c) represents the case when $\lambda=10$. Although the uncertainty of selected samples is high, data points located within a limited range in the feature space are selected in a biased manner.}
    \label{fig:example_duw_expand}
\end{figure*}

In this section, we present qualitative results of sample selection by THC (Fig.~\ref{fig:example_thc_expand}), WPU (Fig.~\ref{fig:example_wpu_expand}), and DUW (Fig.~\ref{fig:example_duw_expand}) on PoseTrack21~\cite{PoseTrack21}.
The experimental settings for each criterion are the same as those in the main paper.

Figs.~\ref{fig:example_thc_expand} and~\ref{fig:example_wpu_expand} demonstrate that our proposed THC and WPU accurately capture incorrect pose estimation results. Furthermore, Fig.~\ref{fig:example_duw_expand} clearly shows that the balance between uncertainty and representativeness changes dynamically with the parameter $\lambda$. This also suggests that when $\lambda=0.01$, we can achieve a selection of uncertain and diverse samples.

\end{document}